\documentclass[12pt,cls]{ucdavisthesis}

\usepackage{etoolbox}
\makeatletter
\patchcmd{\@footnotetext}{\footnotesize}{\scriptsize}{}{}
\makeatother

\usepackage[us,nodayofweek,12hr]{datetime}
\usepackage{graphicx}
\usepackage[round]{natbib}
\usepackage{hyperref}
\usepackage{float}
\usepackage{color,soul}
\usepackage{amsmath}
\usepackage{booktabs}
\usepackage{pbox}
\usepackage{csquotes}
\usepackage{amssymb}
\title          {Ristretto: Hardware-Oriented Approximation of Convolutional Neural Networks}

\author         {Philipp Matthias Gysel}

\authordegrees  {B.S. (Bern University of Applied Sciences, Switzerland) 2012}

\thesis{Master of Science}

\officialmajor  {Electrical and Computer Engineering}

\graduateprogram{Electrical and Computer Engineering}

\degreeyear     {2016}

\degreemonth    {June}

\committee{Professor S. Ghiasi}{Professor J. D. Owens}{Professor V. Akella}{Professor Y. J. Lee}{}

\dedication{\textsl{To my family ...}}

\abstract{
Convolutional neural networks (CNN) have achieved major breakthroughs in recent years. Their performance
in computer vision have matched and in some areas even surpassed human capabilities. Deep neural
networks can capture complex non-linear features; however this ability comes at the cost of
high computational and memory requirements. State-of-art networks require billions of arithmetic
operations and millions of parameters.

To enable embedded devices such as smart phones, Google glasses and monitoring cameras with the
astonishing
power of deep learning, dedicated hardware accelerators can be used to decrease both execution
time and power consumption. In applications where fast connection to the cloud is not guaranteed
or where privacy is important, computation needs to be done locally. Many hardware accelerators
for deep neural networks have been proposed recently.
A first important step of accelerator design is hardware-oriented
approximation of deep networks, which enables energy-efficient inference.

We present Ristretto, a fast and automated framework for CNN approximation. Ristret-to simulates the
hardware arithmetic of a custom hardware accelerator.
The framework reduces the
bit-width of network parameters and outputs of resource-intense layers, which reduces the chip area for multiplication units significantly.
Alternatively, Ristretto can remove the need for multipliers altogether, resulting in an adder-only
arithmetic.
The tool fine-tunes
trimmed networks to achieve high classification accuracy.

Since training of deep neural networks can be time-consuming, Ristretto uses highly optimized
routines which run on the GPU. This enables fast compression of any given network. 

Given a maximum
tolerance of 1\%, Ristretto can successfully condense CaffeNet and SqueezeNet to 8-bit. The code
for Ristretto is available.
}

\acknowledgments{
First and foremost, I want to thank my major advisor Professor Soheil Ghiasi for his guidance, inspiration
and encouragement
that he gave me during my graduate studies. Thanks to him, I had the privilege to
do research in a dynamic research group with excellent students. He provided my with all the ideas,
equipment and
mentorship I needed for writing this thesis.

Second I would like to thank the graduate students at UC Davis who contributed to my research.
I was fortunate to work together with members of the LEPS Group, the Architecture Group as well as the VLSI Computation Lab. Most notably,
Mohammad Motamedi, Terry O'Neill, Dan Fong and Joh Pimentel helped me with advice, technical knowledge and
paper editing. I consider myself extremely lucky that I had their support during
my graduate studies, and I look forward to continuing our friendship in the years to come.
I'm humbled by the opportunity to do research with Mohammad Motamedi, a truly
bright PhD student. Our early joint research projects motivated me to solve
challenging problems and to strive for extraordinary research results.

Third, I am grateful to all members of my thesis committee: Professor John Owens, Venkatesh
Akella, and Yong J. Lee. Professor Owens spurred me on to conduct an in-depth analysis of
related work; additionally he gave me valuable improvement suggestions for my thesis. Early
in this research project, Professor Akella guided me in reading papers on hardware acceleration
of neural networks. Professor
Lee helped me significantly to improve the final version of this document.

Finally I'd like to thank my family for supporting my studies abroad, and especially my girlfriend
Thirza. I am grateful to my family and friends for always motivating
me to pursue my academic goals; without them I would not have come this far.
}

\begin{document}


\setcounter{tocdepth}{1}
\bibliographystyle{myplainnat}

\makeintropages 

\chapter{Introduction}

One of the major competitions in AI and computer vision is the ImageNet Large Scale Visual
Recognition Competition \citep{russakovsky2015imagenet}.
This annually held competition has seen state-of-the-art
image classification accuracies by deep networks such as AlexNet by \cite{krizhevsky2012imagenet},
VGG \citep{Simonyan15}, GoogleNet \citep{szegedy2015going} and ResNet \citep{he2015deep}. All
winners since 2012 have used deep convolutional neural networks. These
networks contain millions of parameters and require billions of
arithmetic operations.

Training of large networks like AlexNet is a very time-consuming process and can take
multiple days or even weeks. The training procedure of these networks is only possible
thanks to recent advancements in graphics processing units. High-end GPUs enable fast
deep learning, thanks to their large throughput and memory capacity. When training
AlexNet with Berkeley's deep learning framework Caffe \citep{jia2014caffe} and Nvidia's
cuDNN \citep{DBLP:journals/corr/ChetlurWVCTCS14}, a
Tesla K-40 GPU can process an image in just 4ms.

While GPUs are an excellent accelerator for deep learning in the cloud, mobile systems are
much more sensitive to energy consumption. In order to deploy deep learning algorithm in
energy-constraint mobile systems, various approaches have been offered to
reduce the computational and memory requirements of convolutional neural networks
(CNNs). Various FPGA-based accelerators \citep{suda2016throughput, Qiu:2016:GDE:2847263.2847265}
have proven that it is possible to use
reconfigurable hardware for end-to-end inference of large CNNs like AlexNet and VGG.
Moreover, we see an increasing amount of ASIC designs for deep CNNs
\citep{isscc_2016_chen_eyeriss, sim201614, han2016eie}.

Before implementing hardware accelerators, a first crucial step consists of condensing the
neural network in question. Various work has been
conducted recently to reduce the computational and memory requirements of neural
networks. However, no open-source project exists which would help a hardware
developer to quickly and automatically determine the best way to reduce the complexity
of a trained neural network. Moreover, too aggressive compression of neural networks
leads to reduction in classification accuracy.

In this thesis we present Ristretto, a framework for automated neural
network approximation. The framework is open source and we hope it will speed up
the development process of energy efficient hardware accelerators. Our framework
focuses on condensing neural networks without adding any computational complexity such
as decompression or sparse matrix multiplication.
Ristretto is a Caffe-based approximation framework and was developed for fast, automated, efficient
and flexible CNN compression for later deployment in hardware accelerators. Ristretto aims at
lowering the area requirements for processing elements,
and lowering the memory footprint which in turn reduces or eliminates off-chip memory communication.

This thesis analyses the resource-requirements of convolutional neural networks. Based
on these findings, 
different approximation strategies are proposed to reduce the resource-intense parts of CNNs.
For all different approximation strategies, we present an in-depth analysis of
the compression vs accuracy trade-off.

Parts of this thesis are based on previous publications with two other authors:
Mohammad Motamedi from the University
of California, Davis and Professor S. Ghiasi from the same university.

\begin{itemize}
\item \textbf{Hardware-oriented Approximation of
Convolutional Neural Networks}, Philipp Gysel, Mohammad Motamedi, and Soheil Ghiasi,
arXiv preprint arXiv: 1604.03168 (2016). \cite{gysel2016hardware}
\item \textbf{PLACID: a Platform for Accelerator Creation for DCNNs}, Mohammad Motamedi,
Philipp Gysel and Soheil Ghiasi, under review. \cite{motamediPLACID2016}
\end{itemize}

\chapter{Convolutional Neural Networks}

Over time, different feature extraction algorithms have been used for image processing tasks.
SIFT \citep{lowe2004distinctive} and HOG (histogram of oriented gradients by
\cite{dalal2005histograms}) were state-of-art for
feature extraction, but they relied
on handcrafted features. Neural networks in contrast can automatically create both high-level
and low-level features. For a long time, deep neural networks were hindered by their computational
complexity. However, advances in both personal computers and general purpose
computing have enable the training of larger networks with more parameters. In 2012, the first
deep convolutional neural network with 8 parameter layers was proposed by
\cite{krizhevsky2012imagenet}.
State-of-the art deep CNNs use a series of convolutional layers which enables them to extract very
high-level features from images. Convolutional neural
networks have proven to overshadow conventional neural networks in complex vision tasks.

\section{Training and Inference}

Convolutional neural networks have two computation phases. The forward propagation is used for
classification, and the backward propagation for training. Like other algorithms in machine
learning, CNNs use a supervised training procedure to find network parameters which yield good
classification accuracy. Throughout this thesis, we use the terms parameters and weights
interchangeably.

\subsection{Forward Propagation}
\label{chap:training}

Input to the forward propagation is the image to classify. The forward path consists of different
layers which process the input data in a chained manner. Deep CNNs use many such layers, the last
of which is used to predict the image class.

\begin{figure}[H]
\includegraphics[width=1.0\linewidth]{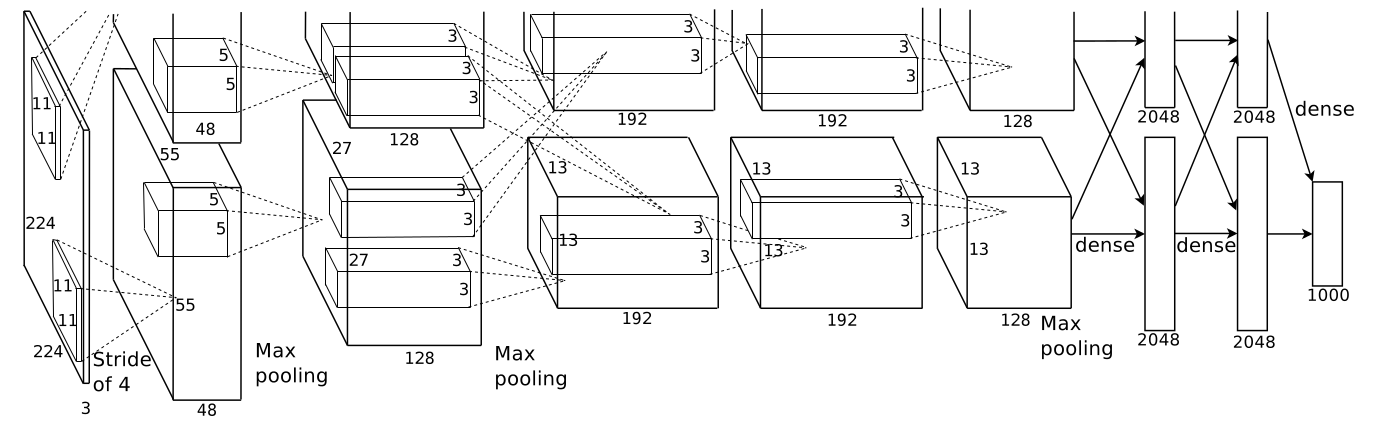}
\centering
\caption[Network architecture of AlexNet]
{Network architecture of AlexNet by \cite{krizhevsky2012imagenet}.}
\label{fig:alexnet}
\end{figure}

A typical CNN architecture termed AlexNet is shown
in Figure \ref{fig:alexnet}. As input to the network, we use an RGB image with dimensions
224$\times$224. The first
five layers in AlexNet are convolutional layers, and the last three are fully connected layers.
Convolutional layers use 2d filters to extract features from the input. The first convolutional
layer generates 2$\times$48 feature maps, each of which represents the presence or absence of a
low-level
feature in the input image. To reduce the spatial dimension of feature maps and to add
translation-invariance, pooling layers
are used which do sub-sampling. Moreover, a non-linear layer is added which enables the network 
to learn non-linear features.

The last convolutional layer creates 2$\times$128 feature maps with spatial dimension
13$\times$13.
This
layer is followed by three dense layers, which produce the weighted sum of inputs. The last layer
has 1000 output nodes, which are the predictions for the 1000 image classes.

Forward propagation depends on network parameters in two ways. First, the convolutional layers
rely on feature filters. Second, the fully connected layers contain many parameters, each
of which serves as weighted connection between a specific input and output node. These parameters
are learned during the training procedure.

\subsection{Backward Propagation}

Training of a deep CNN requires thousands or even millions of labeled images. The network
is exposed to these images, and the network parameters are gradually updated to
make up for prediction errors. The purpose of backward propagation is to find the error
gradient with respect to each network parameter. In a later step this error gradient is used
for a parameter update.

Training is done in batches of images. Several images are run through the network in forward path.
Denoting $x$ as the network input, $w$ as the network parameters and $f$ as the overall CNN
function, the network output is given by $z'=f(x,w)$. Since all images are labeled, the desired
network output $z$ is known. Given many pairs of images and ground-truth $(x_1,z_1),...,(x_n,z_n)$,
we define a loss function $l(z,z')$ which denotes the penalty for predicting $z'$ instead of
$z$.

We average the loss for a batch of images, and update the parameters according to the formula below:

\begin{equation}
w^{t+1}=w^{t}-\alpha \cdot \frac{\delta l}{\delta w}(w^t)
\end{equation}

This formula requires us to calculate the network output error w.r.t. each parameter. The
above parameter update is called stochastic gradient descent, which relies on a learning
rate $\alpha$. There actually exist many optimizations for this parameter update such as
Nesterov momentum as explained by \cite{bengio2013advances} or Adam rule \citep{kingma2014adam}.
All these learning rules require the error gradient w.r.t each network parameter.
Moreover these optimization procedures all work in batch mode, i.e., a batch of images
(e.g.~256 images) is run through the network
and the parameter update is based on the average error gradient. The error surface of neural
networks is non-convex. Nevertheless, batch-based learning rules can avoid local minima by using
many different training examples.

For computation of the error gradient w.r.t the network parameters, we first compute
the error gradient with respect to each layer output, starting with the second last layer and
back propagation to the second layer. In a second step, the gradients w.r.t. the layer
outputs can be used to compute the gradients w.r.t. the network parameters, using the
chain rule for derivatives.

\section{Layer Types}
Deep convolutional neural networks process the input data layer by layer. Each layer has a specific
purpose as explained in the following paragraphs.

\paragraph{Convolutional layer:}

\begin{figure}[H]
\includegraphics[width=0.5\linewidth]{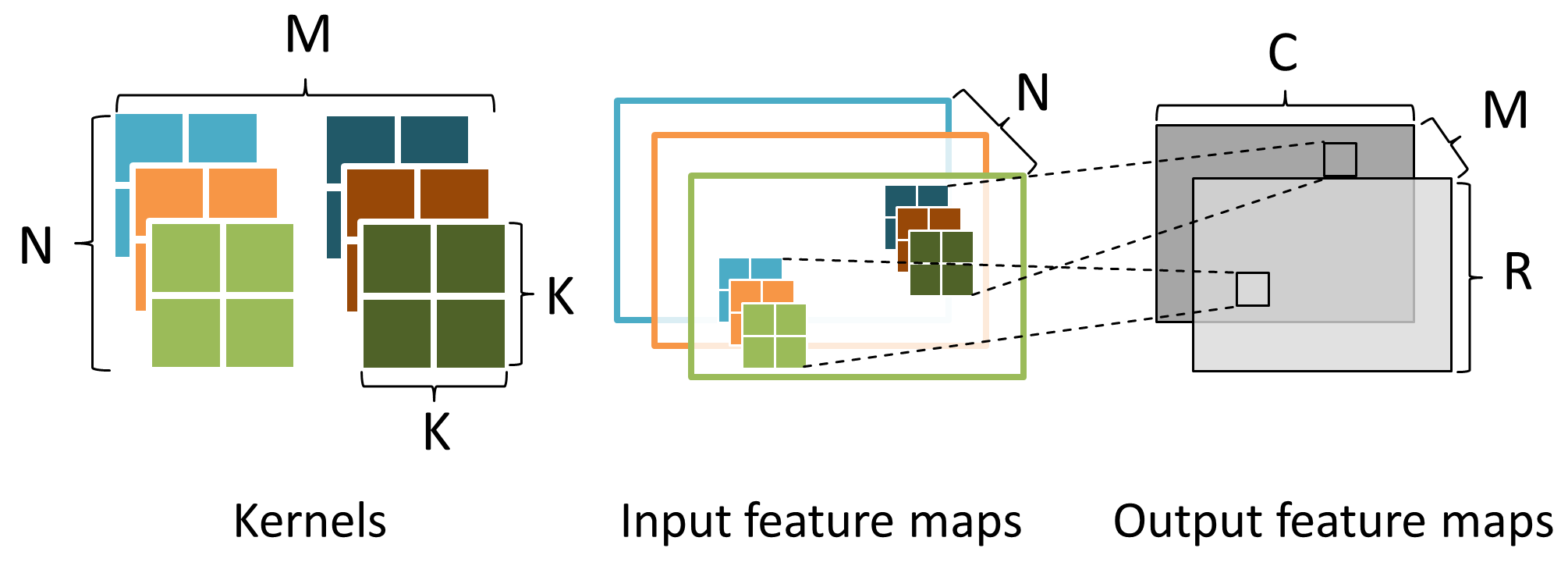}
\centering
\caption[Convolution between input feature maps and filters]
{Convolution is the result of extracting 2d features from multiple feature maps. Image credit:
\cite{motamediPLACID2016}.}
\label{fig:conv}
\end{figure}

\begin{figure}[H]
\includegraphics[width=0.4\linewidth]{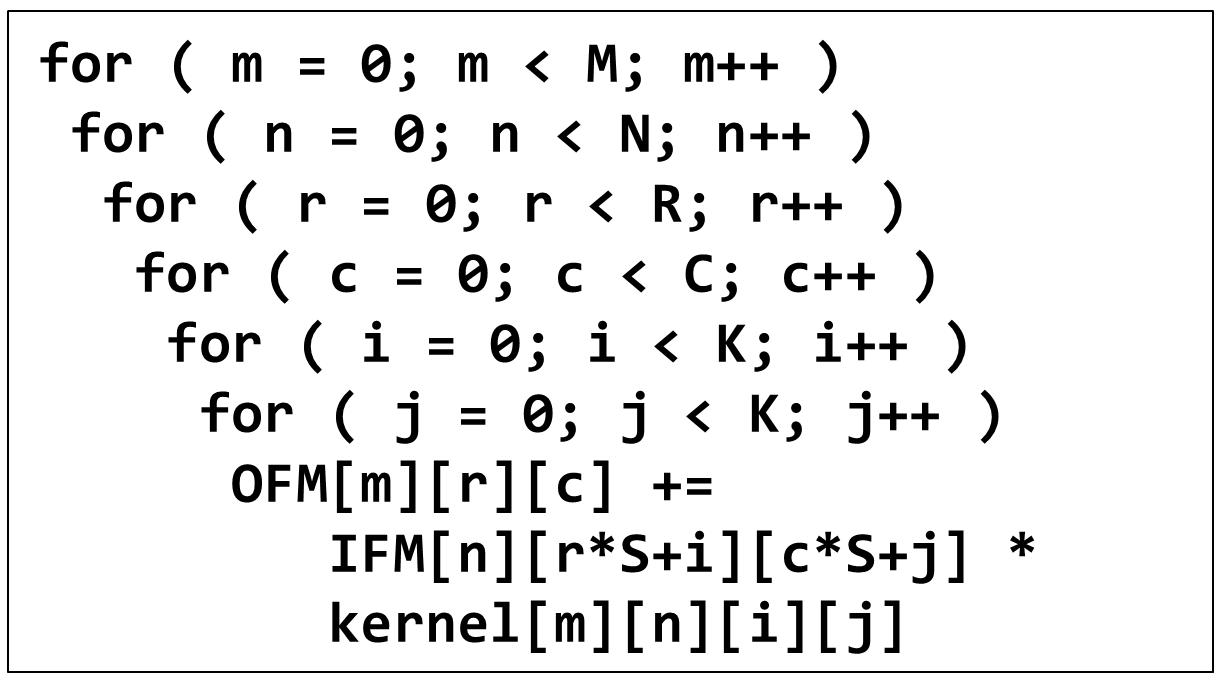}
\centering
\caption[Pseudo-code for convolutional layer]
{Pseudo-code for convolutional layer. Image credit: \cite{motamediPLACID2016}.}
\label{fig:conv_code}
\end{figure}

This layer type is used for feature extraction. Deep CNNs have many convolutional layers; AlexNet
for example has five layers of this type.
The feature extraction is done by a series of $L$ convolutional
layers. Each layer uses 2d kernel filters which extract features
from input feature maps (IFM). The result of multiple
feature extractions is summed up to form one output feature
map (OFM). This process is shown in Figure \ref{fig:conv}, where two
filter banks, each with three kernels, are used to generate 2
output feature maps. The number of input and output feature
maps is denoted by $N$ and $M$, and the size
of one output feature map is $R{\times}C$. One kernel has dimension $K{\times}K$,
and a layer has $N{\times}M$ of these kernels. The feature extraction
consists of a series of multiplication-accumulation
(MAC) operations, as shown in Figure \ref{fig:conv_code}. Each output
pixel is the sum of the 2d convolutions between the input
feature maps and the respective kernels. To generate the
neighboring output pixels, the kernel stack is slided across
the spacial dimension by stride $S$.

The time and space complexity of convolutional layers is given in Equations \ref{eq:conv_time}
and \ref{eq:conv_mem}. Assuming a stride $S$ of 1, the computational complexity is $R \cdot C$
times larger than the number of parameters. It is for this reason that the computational
bottleneck of deep CNNs comes from convolutional layers.

\begin{equation}
\label{eq:conv_time}
runtime=\mathcal{O}(RCMNK^2)
\end{equation}

\begin{equation}
\label{eq:conv_mem}
parameter \; size=\mathcal{O}(MNK^2)
\end{equation}

\paragraph{Fully connected layer:}

Fully connected layers serve the same purpose as convolutional layers, namely feature extraction.
Fully connected layers, alternatively termed dense layers, build the weighted sums of their input.
Thus all input nodes are connected to all output nodes, which requires a relatively large amount
of network parameters. Fully connected layers are the basic building block of \textit{classic
neural networks}, which are normally a concatenation of dense layers. In \textit{convolutional
neural networks},
the first layers are normally convolutional layers, and only one or two layer at the very end
are dense.

The mathematical function of a fully connected layer is a simple matrix-vector product.
The layer output nodes $\bf{z}$ depend on the input vector $\bf{x}$, 
the parameter-matrix $W$ and the bias $\bf{b}$ (Equation \ref{eq:fc}). Denoting $N=|\bf{x}|$ as
the number of input nodes and $M=|\bf{z}|$ as the number of outputs, the time and space complexity
is given in Equations \ref{eq:fc_time} and \ref{eq:fc_space}. For layers with many input
nodes, the parameter size usually makes up for a large part of the overall network size.

\begin{figure}[H]
\includegraphics[width=0.6\linewidth]{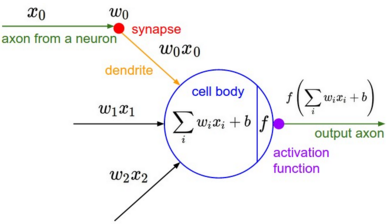}
\centering
\caption[Fully connected layer with activation]
{Fully connected layer followed by an activation function. Image credit:
\cite{karpathy2016stanford}.}
\label{fig:fc1}
\end{figure}

Most fully connected layers are followed by a non-linear activation function $f$.
Figure \ref{fig:fc1}
depicts a fully connected layer followed by an activation layer.

\begin{equation}
\label{eq:fc}
\bf{z} = W \cdot \bf{x} + \bf{b}
\end{equation}

\begin{equation}
\label{eq:fc_time}
runtime = \mathcal{O}(NM)
\end{equation}

\begin{equation}
\label{eq:fc_space}
parameter \; size = \mathcal{O}(NM)
\end{equation}

\paragraph{Rectified Linear Unit (ReLU):}

To enable neural networks to capture non-linear relations, several non-linear layers are
inserted into the network architecture. Traditionally we used the Sigmoid or hyperbolic
tangent function for this purpose. However, those classic non-linear functions have several
disadvantages. First, their gradient becomes very small for large values, which means
the error gradient during training vanishes. Moreover these two non-linear functions are
relatively costly in terms of computing power.

As alternative, nearly all state-of-art CNNs use Rectified Linear Units (ReLU). This
activation function was first proposed by \cite{nair2010rectified} for Restricted Boltzmann
Machines. The work by \cite{krizhevsky2012imagenet} was the first to apply this simplified
activation to a deep neural networks. Deep networks trained with this activation function
converge much faster.

The function of a ReLU layer maps negative values to zero: $f(x)=max(0,x)$.

\paragraph{Normalization layers:}

Local Response Normalization (LRN) layers serve the purpose of normalization across feature
maps or across the spatial dimension. There are generally two type of LRN layers: LRN
across feature maps and LRN within feature maps. The first type of normalization serves the
purpose of normalizing neurons in the same spatial position, but different feature maps. This
creates competition between neurons generated by different kernels and improves the top-1
accuracy of AlexNet \citep{krizhevsky2012imagenet} by 1.4\%. The exact mathematical formulation
of LRN across channels can be found in the same paper.

Many state-of-art CNNs use LRN layers to increase accuracy \citep{krizhevsky2012imagenet,
szegedy2015going}. One notable exception is the network by \cite{Simonyan15} which performs
very well without any kind of normalization.

Most recently, a new normalization strategy termed Batch Normalization (BN) was proposed
\citep{ioffe2015batch}. This strategy was adopted by the most recent winner of the ILSVRC
competition \citep{russakovsky2015imagenet}.

While both normalization strategies help for faster convergence and better prediction accuracy,
they also add computational overhead, especially batch normalization. Unfortunately these
normalization layers require a very large dynamic range for intermediate values. In AlexNet
for example, the intermediate values of LRN layers are $2^{14}$ times larger than any intermediate
value from another layer. For this reason this thesis assumes LRN and BN layers are to be
implemented in floating point, and we concentrate on the approximation of other layer types.
Notice that previous work by \cite{suda2016throughput} chose 32-bit floating point for
FPGA-based LRN layers.

\paragraph{Pooling:}

\begin{figure}[H]
\includegraphics[width=0.6\linewidth]{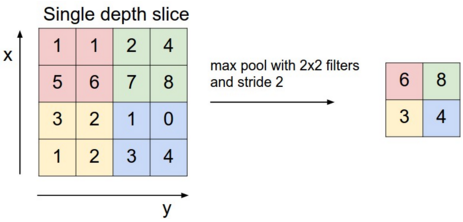}
\centering
\caption[Pooling layer]
{Max pooling layer with 2$\times$2 filter size. Image credit: \cite{karpathy2016stanford}.}
\label{fig:pooling}
\end{figure}

Pooling layers are normally used between successive convolutional layers in CNNs. They can
be considered as sub-sampling functions. The purpose of pooling layers is to reduce the spatial
dimension of feature maps and encode translation invariance. By reducing the feature map
dimensions, we also reduce the parameter size and computational complexity, and thus lower chances
for over-fitting.

The most commonly used pooling type in CNNs is MAX pooling, although other types such as average 
pooling and L2-norm pooling exist. MAX pooling does not required any arithmetic operation except
comparisons. Since this layer type is cheap in terms of computation and parameter size, this
thesis leaves this layer type as is and we perform no approximation.

An example of a MAX pooling operation is shown in Figure \ref{fig:pooling}, where 2x2 kernels
are used to extract the maximum value. The result is stored as output on the right side of
the Figure. Pooling layers are associated with a kernel size and a stride. The stride indicates
how many pixels are skipped between pooling operations. Notice that in the above example, the
data size is reduced by 75\%.

\section{Applications}

Deep convolutional neural networks have pushed the limits of artificial intelligence in a wide
range of
applications. Recent winners \citep{krizhevsky2012imagenet, Simonyan15, szegedy2015going,
he2015deep} of ImageNet \citep{russakovsky2015imagenet} competition have continuously
improved machine's abilities in image classification. The most recent winners of this
competition even outperform human vision. Besides image classification, deep networks
show state-of-art performance in object detection \citep{girshick2015fast} as well as speech
recognition \citep{hinton2012deep}. Other applications include playing games
\citep{silver2016mastering}, as well as art \citep{gatys2015neural}. Notice that artificial
neural networks as well as recurrent neural networks could potentially be approximated in
a similar way as we do in this thesis.

\section{Computational Complexity and Memory Requirements}

The complexity of deep CNNs can be split into two parts. First, the convolutional layers
contain more than 90\% of the required arithmetic operations. The second 
resource-intense layer type are fully connected layers, which contain over 90\% of the network
parameters. An energy-efficient accelerator for CNNs needs to 1) offer a large enough
computational throughput and 2) offer a memory-bandwidth that can keep the processing
elements busy.

\begin{figure}[H]
\includegraphics[width=0.7\linewidth]{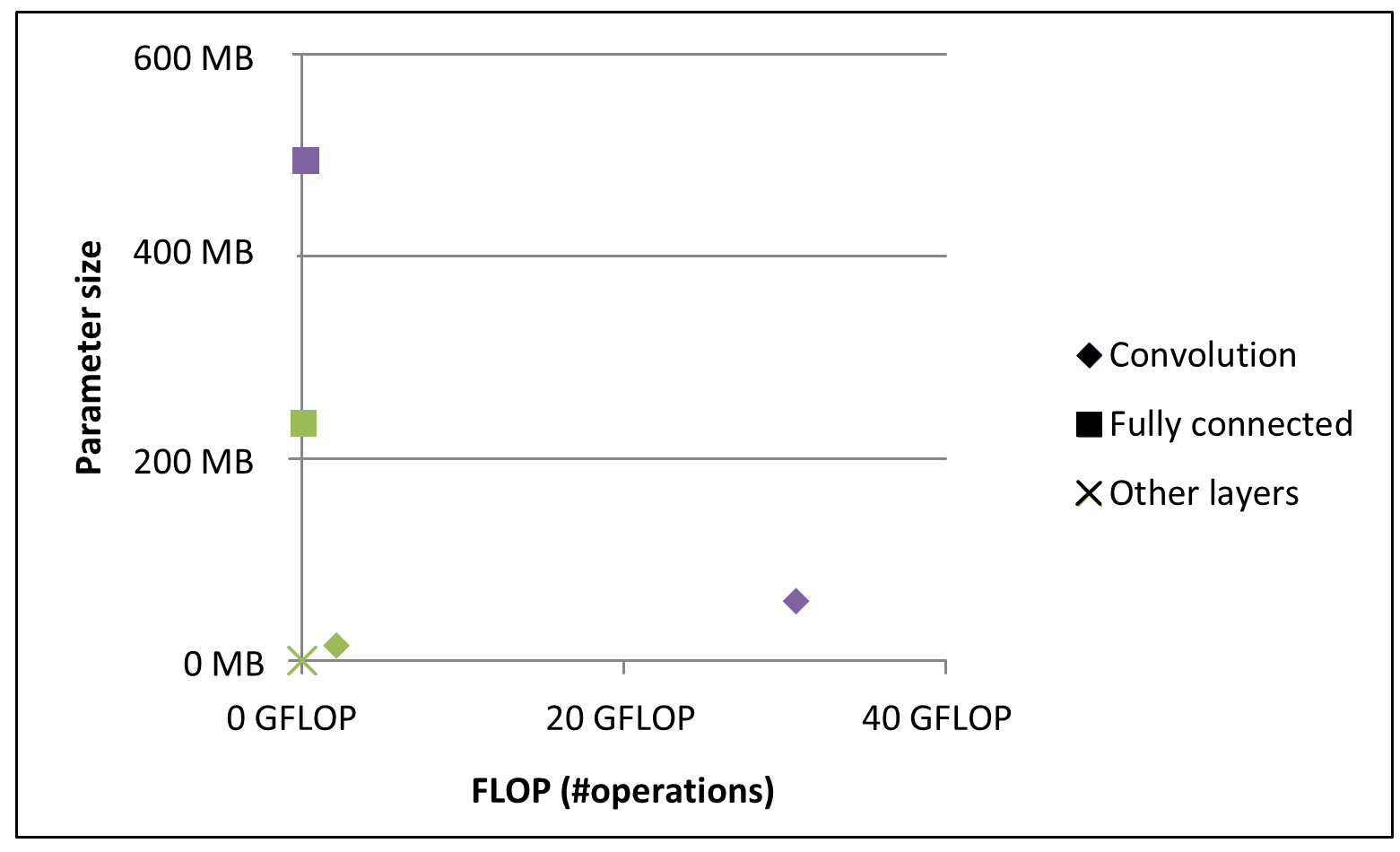}
\centering
\caption[Parameter size and arithmetic operations in CaffeNet and VGG-16]
{Parameters and arithmetic operations in CNNs. Data from CaffeNet is colored green, data from
VGG-16 network is colored violet.}
\label{fig:ops_size_cnn}
\end{figure}

CaffeNet is the Caffe-version of AlexNet by \cite{krizhevsky2012imagenet}. CaffeNet was developed
for the ImageNet data set, which has 1000 image classes. Figure \ref{fig:ops_size_cnn} shows
the required arithmetic operations and parameters size of AlexNet by layer type. The major
part of arithmetic operations comes from convolutional layers: this layer type requires
a total of 2.15 G operations. The arithmetic operations in all other layers sum up to 117~M
operations.
The parameter size of CaffeNet is 250 MB, of which 235 MB comes from fully connected layers.

The same trend can be observed for the 16-layer version of VGG by \cite{Simonyan15}: Extracting
features in convolutional layers is computation-intense, while fully connected layers are
memory-intense. Since most computation and memory requirements come from fully connected and
convolutional layers, this thesis concentrates on approximating these two layer types only.

\section{ImageNet Competition}

The ILSVRC competition \citep{russakovsky2015imagenet} is a large scale image classification
and detection challenge which has been held annually since 2010. More than fifty institutions
have participated in the challenge, among them companies like Google, Microsoft, Qualcomm,
as well as various universities. In its first year the competition consisted of an image
classification challenge only. In the meantime, different new challenges were introduced such
as object
detection and localization in images, object detection from video as well as scene
classification.

For training of the classifier, ILSVRC provides the ImageNet data set, which consisted originally
of 1000 image categories and a total of 1.2 million images. In the meantime, this data base
has been significantly expanded, and currently holds over 14 million labeled images.

A common performance measure for deep CNNs is their classification accuracy on the ILSVRC 2012
data set. For this challenge, the classifier is trained on a
\textit{training data set}, and tested by the researcher on a \textit{validation data set}. For
the official score in the ILSVRC competition, a private \textit{test data set} is used which
is not available to the public. Most researchers give their performance numbers in top-1 and top-5
measure. The top-1 measure gives the percentage of images that were classified correctly on either
the validation or test set. Since the ImageNet data set has very fine-grained image classes which
are sometimes even hard for humans to distinguish, most researchers prefer to use the top-5 measure.
Here, an image is considered as classified correctly if one of the top-5 predictions is correct.
In the remainder of this thesis, we will concentrate on top-1 accuracy, since this captures better
how CNN approximation affects network accuracy.

\subsection{Network Size vs Accuracy}

On one hand, recent network architectures indicate that deeper networks perform better.
The winner of 2014's localization challenge \citep{Simonyan15} experimented with different
depths for
their network. Going from 11 parameter layers to 19 improved their top-1 classification
for over 4\%. Another experiment by the winner of 2015 \citep{he2015deep} used very deep networks.
Their network architecture improves by over 2\% when expanding the net from 34 to 152
layers.

On the other hand, some research shows that even relatively small networks can achieve
good classification performance. In the classification challenge of 2014, GoogLeNet
\citep{szegedy2015going}
outperformed VGG \citep{Simonyan15} with a network capacity that was over 19X smaller. The 
GoogLeNet network is based on the inception idea described in section \ref{chap:inception}.
A newer network architecture by \cite{SqueezeNet} uses an adapted inception concept with smaller
convolutional kernels. Their small, yet quite accurate network termed SqueezeNet was developed
specifically for mobile devices. SqueezeNet has the
accuracy of AlexNet \citep{krizhevsky2012imagenet}, but contains 50X fewer parameters.
SqueezeNet avoids fully connected layers,
which drastically reduces the parameter size to below 5 MB.

\begin{figure}[H]
\includegraphics[width=0.7\linewidth]{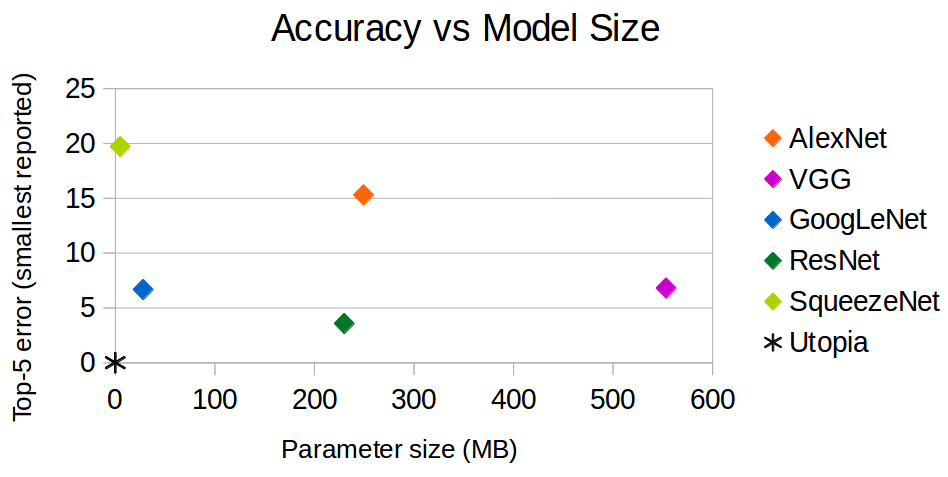}
\centering
\caption[ImageNet networks: accuracy vs size]{ImageNet networks: accuracy vs size.}
\label{fig:cnns_acc_vs_size}
\end{figure}

In summary, there is a delicate trade-off between network size and accuracy (see Figure
\ref{fig:cnns_acc_vs_size}). GoogLeNet finds a good balance for reducing both classification
error and network size. This network by Google outperforms AlexNet and VGG in both aspects.
SqueezeNet is the smallest architecture, but its accuracy is not
outstanding. ResNet has the best accuracy, but it is also the largest network both in terms
of parameter layers and parameter size.

\subsection{Inception Idea}
\label{chap:inception}
\begin{figure}[H]
\includegraphics[width=0.8\linewidth]{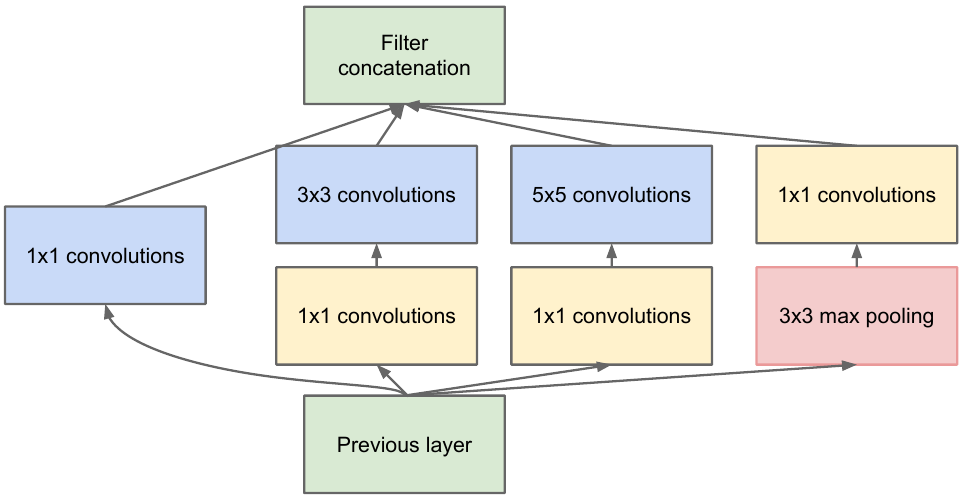}
\centering
\caption[Inception architecture]{Inception architecture used by GoogLeNet. Image credit: \cite{szegedy2015going}.}
\label{fig:inception}
\end{figure}

The inception idea is a concept proposed by \cite{szegedy2015going} which was used to
build the GoogLeNet architecture. GoogLeNet was developed with the goal of achieving
high classification accuracy on the ImageNet data set with a budget of 1.5 billion MAC operations
for inference. The authors avoid sparse representations and chose to use readily available
dense components. The GoogLeNet
architecture contains several replicas of the inception module shown in Figure \ref{fig:inception}.
An inception module contains 1$\times$1, 3$\times$3 and 5$\times$5 convolution kernels. In order
to reduce the number of feature maps and thus the computational complexity, 1$\times$1
convolutions
are added in front of the expensive 3$\times$3 and 5$\times$5 convolutions. The inception
module also
contains
an alternative max pooling path.

\section{Neural Networks With Limited Numerical Precision}
\label{chap:nn_limited_precision}

This section focuses on limited precision arithmetic for convolutional neural networks.
Most deep learning frameworks \citep{jia2014caffe, 2016arXiv160502688short,
tensorflow2015-whitepaper} use 32-bit or 64-bit floating point for CNN training and
inference. However, it has be shown \citep{du2015leveraging} that CNNs have a relatively high
error resilience;
moreover CNNs can be trained in a discrete parameter space. In the following subsections, we
describe the process of quantizing a full precision network to limited precision numbers.
Additionally we introduce different rounding schemes for the quantization step, and we describe
different options for optimizing the classification accuracy of a limited precision network.

\subsection{Quantization}
\label{chap:quantization}

\begin{figure}[H]
\includegraphics[width=0.8\linewidth]{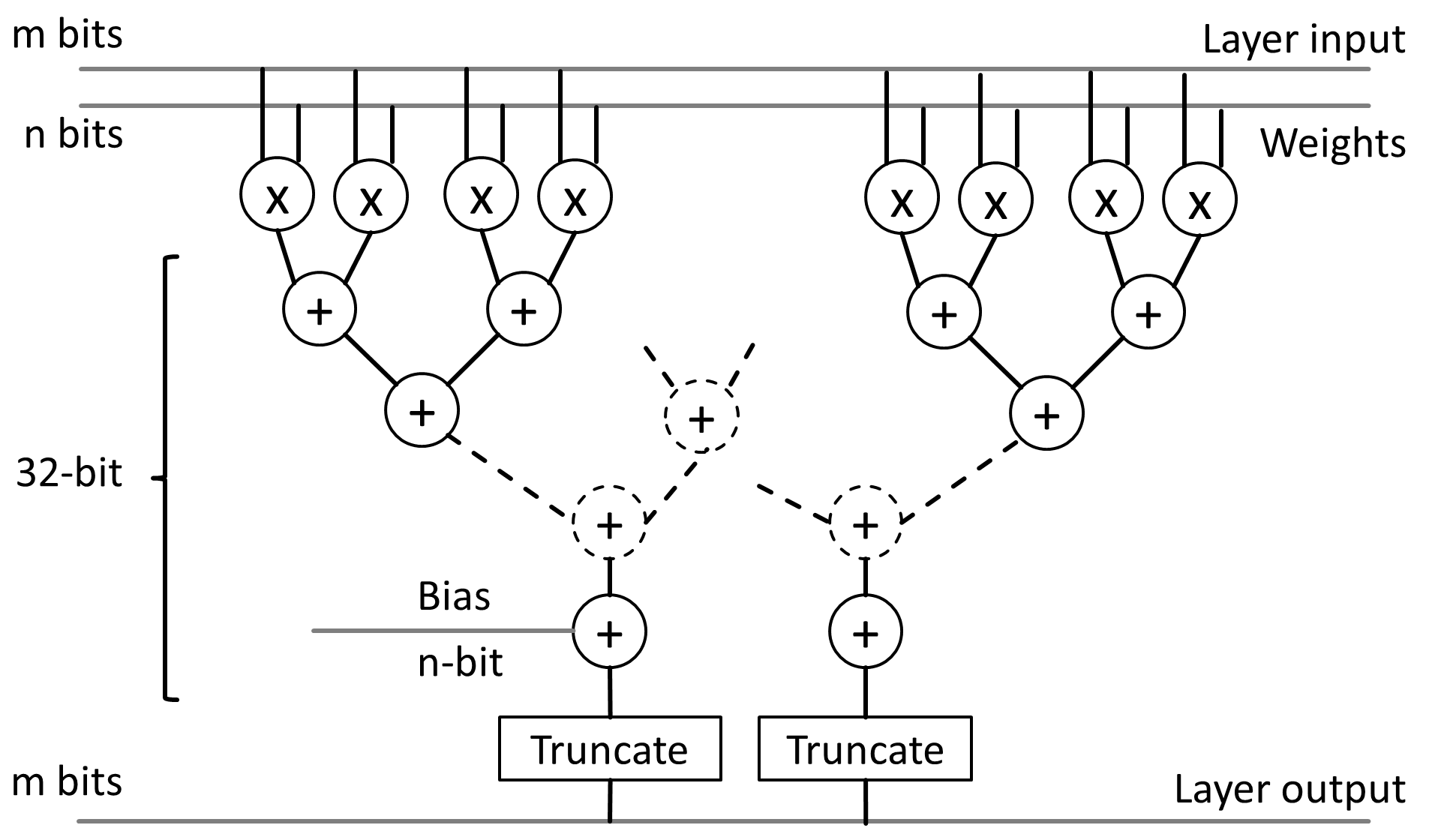}
\centering
\caption[Data path with limited numerical precision]{
Simulated data path for convolutional and fully connected layers. The layer inputs, layer outputs
and layer parameters are discrete-valued numbers.
}
\label{fig:data_flow_any_format_hw}
\end{figure}

As discussed previously, the goal of this
thesis is to provide a framework for approximating the forward path of any given CNN. For this
purpose we compress the number format in convolutional and fully connected layers. These two
layer types, which are the most resource-demanding part of a deep network, require the same
arithmetic operations, namely a series of multiplication-and-accumulation (MAC). In this
thesis we simulate the arithmetic of a hardware accelerator. The
simulated data path is shown in Figure \ref{fig:data_flow_any_format_hw}.

The difference between this simulated data path and the original full precision data path is the
quantization step of weights, layer inputs, and layer outputs. Therefore the condensed networks will suffer
from quantization errors, which can affect the network accuracy.

In this thesis we propose a framework which can approximate 32-bit floating point networks by
condensed ones which use quantized values. In order to simulate a condensed layer, given
a full precision reference network, the following three steps are required:

\begin{itemize}
\singlespacing
 \item Quantization of the layer input and weights to reduced precision format (using $m$ and
 $n$ bits for number representation, respectively)
 \item Perform the MAC operations using the quantized values
 \item The final result is again quantized
\end{itemize}

\begin{figure}[H]
\includegraphics[width=0.5\linewidth]{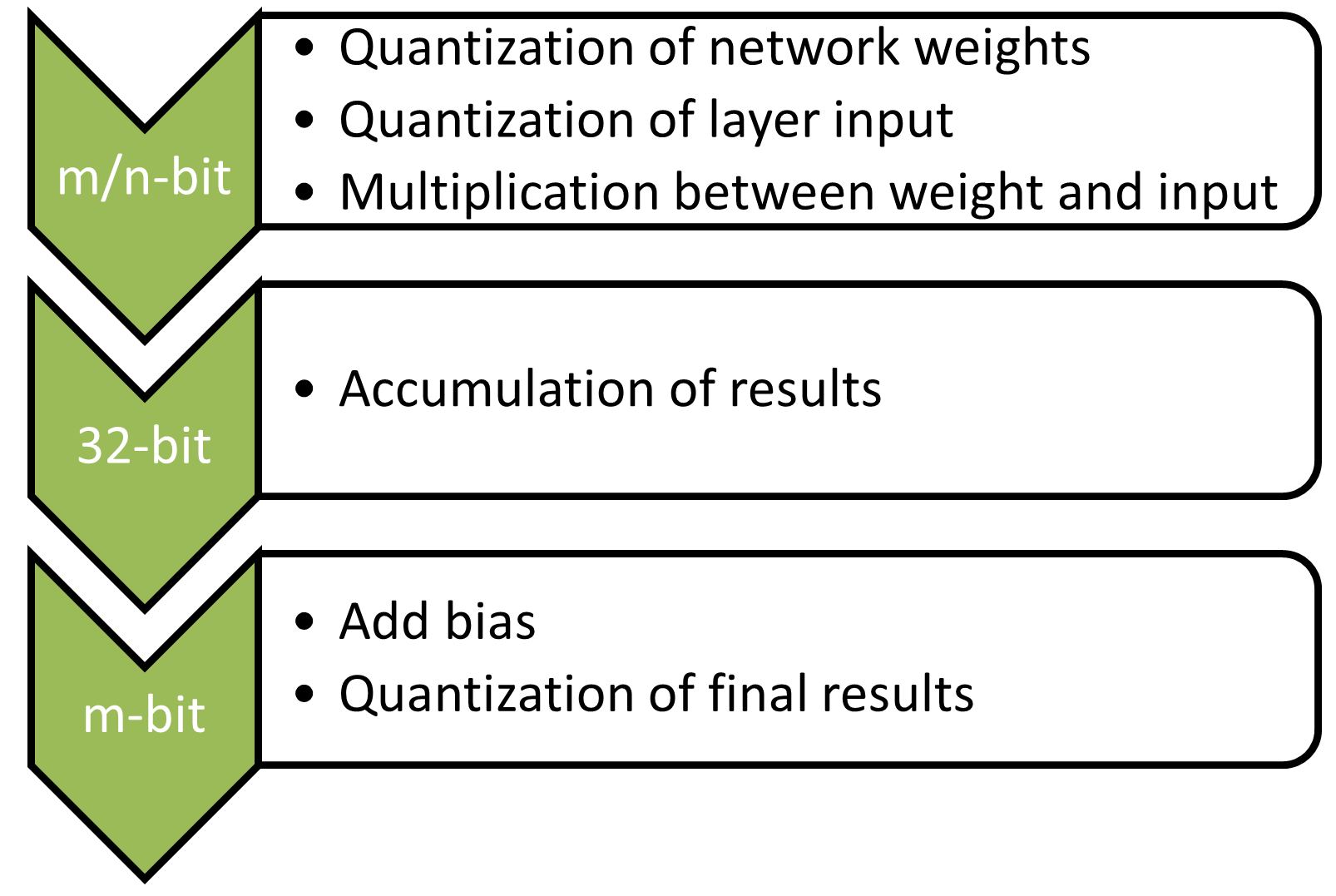}
\centering
\caption[Quantized layer]{Our quantization framework performs these steps to simulate the hardware
arithmetic.}
\label{fig:layer_quantization}
\end{figure}

These steps are summarized in Figure \ref{fig:layer_quantization}.

\subsubsection{Data path in hardware}
\label{chap:approximation_ristretto}

In Figure \ref{fig:data_flow_any_format_hw}, the layer input values and layer weights serve as
input to the multiplication units. To leverage
the inherent sources of parallelism, a hardware accelerator will use many of these units in
parallel. Each multiplication unit gets one layer input value and one network weight per
computation round. The different results are accumulated in an adder tree, and the final sum
is the layer output. Notice that some of the layer input and weight values are actually reused
for different multipliers, depending on the exact characteristics of the layer in question.

To reduce the number of bits required for number representation, our approximation framework
quantizes both the
layer inputs and weights. The resulting values can be represented using significantly fewer bits.
As a result, the multiplication units require less area. In order to simplify simulation of 
hardware, our framework uses 32-bit floating point for accumulation. To achieve the same result
in a hardware accelerator, the adder tree should use 32-bit.
Adders are much cheaper in terms of area and power, compared to multipliers. Thus it is acceptable
to use more bits for number representation in the adder tree.

As a last step, the layer output is quantized to limited-precision format. This reduces the
memory requirements, especially if results need to be written back to off-chip memory.

\subsection{Rounding Schemes}

Different rounding schemes can be used for value quantization.

\paragraph{Round nearest even:}
Round-nearest-even
is an unbiased
scheme which rounds to the nearest discrete value. 
Denoting $\epsilon$ as the quantization step size
and $\lfloor x \rfloor$ as the largest quantization value less or equal to $x$, \cite{gupta2015deep}
define round nearest as follows:
\begin{equation}
  round(x)=
  \begin{cases}
   \lfloor x \rfloor, & \text{if } \lfloor x \rfloor \leq x \leq x+\frac{\epsilon}{2} \\
   \lfloor x \rfloor + \epsilon, & \text{if } \lfloor x \rfloor + \frac{\epsilon}{2} < x \leq x+\epsilon \\
  \end{cases}
\end{equation}
As round-nearest-even is deterministic,
we chose this rounding scheme for inference, i.e., at test time all the parameters are rounded
deterministically, and so are the layer outputs.

\paragraph{Round stochastic:}

Another rounding scheme termed stochastic rounding was used by \cite{gupta2015deep} for
the weight updates of 16-bit neural networks. \cite{gupta2015deep} define stochastic rounding as
follows:

\begin{equation}
  round(x)=
  \begin{cases}
   \lfloor x \rfloor, & \text{w.p. } 1-\frac{x-\lfloor x \rfloor}{\epsilon} \\
   \lfloor x \rfloor + \epsilon, & \text{w.p. } \frac{x-\lfloor x \rfloor}{\epsilon} \\
  \end{cases}
\end{equation}

Stochastic rounding adds randomness to the
quantization
procedure, which can have an averaging effect during training. We chose to use this rounding
scheme when quantizing network parameters during fine-tuning, as explained later in the next
subsection. Stochastic rounding has an expected rounding error of zero, i.e.
$\mathbb{E}(round(x))=0$.

\subsection{Optimization in Discrete Parameter Space}

Training of neural networks can be seen as an optimization problem, where the goal is to find
the optimal set of parameters which minimizes the classification error for a given set of images.
A practical solution to this problem is to use stochastic gradient descent, as explained
in subsection \ref{chap:training}. In the traditional setting of 64-bit floating point training,
this optimization is a continuous problem with a smooth error surface.
The error surface of neural networks depends on
its input and its current parameters. For quantized networks, this error surface becomes
discrete. This new optimization problem -- where the goal is to find an optimal set of discrete
valued parameters --
is an NP-hard problem. One approach to find a good set of discrete parameters
is to train in limited precision `from
scratch'. In this approach, we would train the network with quantized parameters right from the
start. All the weight updates would be discrete.\\
We chose to use another approach: Our framework first trains a network in the continuous
domain, then quantizes the parameters, and finally fine-tunes in discrete
parameter space. This way we can fully leverage pre-trained networks which saves considerable
amount of time.

During this retraining procedure, the network learns
how to classify images with limited precision parameters. Since the network weights can
only have discrete values, the main challenge consists in the weight update. We adopt the
idea of previous work by \cite{courbariaux2015binaryconnect} which uses full precision shadow
weights. Small weight
updates $\Delta w$ are applied to the full precision weights $w$, whereas the discrete weights
$w'$
are sampled from the full precision weights. The sampling during fine-tuning is done
with stochastic rounding. For more details on the fine-tuning procedure of quantized networks,
please refer to section \ref{chap:fine-tuning}.
\chapter{Related Work}

In the first section of this chapter, we review different network approximation techniques.
In the second part, we describe related work in hardware accelerator design.

\section{Network Approximation}

\subsection{Fixed Point Approximation}

Various solutions have been offered to reduce the resource-requirement of CNNs. Traditionally
neural networks are trained in 32-bit floating point. However fixed
point arithmetic is less resource hungry than floating point arithmetic. Moreover, it has
been shown that fixed point arithmetic is adequate for neural network computation. This
observation has been leveraged recently to condense deep CNNs. \cite{gupta2015deep} show
that networks on datasets like CIFAR-10 (10 images classes) can be trained in 16-bit.
Further trimming of the same network uses as low as 7-bit multipliers \citep{courbariaux2014low}.
Another approach by \cite{courbariaux2015binaryconnect} uses only binary weights, again on the
same network. A similar proposal represents the weights of the same network with +1, 0 and
-1 values \citep{sung2015resiliency}. While these proposed fixed point schemes work well with
small networks, only limited work has been done on large CNNs like AlexNet.

\subsection{Network Pruning and Shared Weights}

Off-chip memory access makes up for a significant part of the total energy budget of any
data-intense application. As deep CNNs have typically more than 10 MB of parameters,
an important step is to compress the size of the network parameters. The deep
compression pipeline proposed by \cite{han2015deep} addresses this problem. The authors
achieve a
network parameter compression rate of up to 49X for deep CNNs using a three-step
pipeline. In a first step, the 'unimportant' connections of a trained network are removed.
The resulting sparse network is then retrained to regain its classification accuracy, and the
pruning step is repeated. After some iterations of pruning and fine tuning, the remaining
parameters are clustered together to form shared weights. These shared weights are again
fine-tuned to find optimal centroids. In a last step, a lossless data compression scheme
(Huffman Coding) is applied to the final weights.

\subsection{Binary Networks}

Since memory access has a relatively high energy cost, it is desirable to reduce the
network parameter size. This motivated BinaryConnect \citep{courbariaux2015binaryconnect},
a work which represents
weights in binary format, rather than in traditional 32-bit floating point. This approach
reduces parameter size by factor 32X and removes the need of multiplications in the
forward path. BinaryConnect achieves near-state-of-art performance on 10-class datasets
(MNIST, CIFAR-10, SVHN).

A later work by \cite{lin2015neural} takes this idea a step further by turning
multiplications in the backward propagation into bit shifts. Layer activations are
approximated by integer power of 2 numbers, while error gradients are retained in full
precision. This proposal significantly reduces the hardware requirements for accelerators.

Combining the two previous ideas, `Binarized Neural Network' \citep{DBLP:journals/corr/CourbariauxB16}
uses binary weights
and layer
activations. These numbers are constraint to +1 and -1 for both forward and backward
propagation. Convolutional neural networks mainly consist of multiply-accumulate
operations. For a binarized network, these operations are replaced by binary XNOR and binary
count. To improve training results, the proposed method uses a bit-shift-based batch normalization
as well as a shift-based parameter update.

Finally the work of \cite{rastegari2016xnor} applies the idea of binary networks to ImageNet
data set. The
three previously mentioned approaches work well with small networks; however they
show limited performance on large networks for the ImageNet data set. The work by
\cite{rastegari2016xnor} proposes two
network architectures. Both are based on AlexNet \citep{krizhevsky2012imagenet} and use
different degrees of
binarization. First, the proposed Binary-Weights-Network shows a speedup of 2X for
CPU implementation and achieves an accuracy within 3\% of AlexNet. Second, XNOR-
Net has binary weights and layer outputs. XNOR-Net turns most MAC operations into
binary XNOR and bit count, however at a relatively high accuracy drop (12.4\%).

\section{Accelerators}

Different accelerator platforms have been used to accelerate CNN inference. In what follows we
review proposed accelerators on GPUs, FPGAs and ASICs.

\subsection{GPU Acceleration}

Given the high throughput and memory bandwidth of today's GPUs, different research
has focused on accelerating GPU-based inference of neural networks. A
proposal by \cite{denton2014exploiting} uses clustered filters and low rank approximation. They
achieve a speedup of 2X for convolutional layers of AlexNet, compared to a non-optimized
GPU implementation. Another work by \cite{mathieu2013fast} achieves better
results by replacing convolution through FFT. Finally the neural network compression
pipeline proposed by \cite{han2015deep} uses pruning and weight-sharing. When this compression is
applied to dense layers of AlexNet, forward propagation is 4X faster and 3X more energy
efficient. In their later paper \citep{han2016eie}, they show a Titan X based GPU implementation
with a throughput of 3.23 TOPs.

High-end GPUs require a lot of energy. As a case in point, Nvidia's Tesla K-40 has an
average power consumption of 235 W when running DGEMM. This motivated
researchers to consider accelerator platforms with smaller power budgets.

\subsection{FPGA-based Accelerators}

Field programmable gate arrays (FPGA) can offer high throughput per power. FPGA-based
accelerators have a shorter development time than ASICs, however they can't match the
throughput of GPUs. Different FPGA-based accelerators for neural networks have been
proposed. An approach by \cite{zhang2015optimizing} uses Vivado HLS to accelerate the
convolutional layers of
AlexNet. Their floating point implementation achieves a throughput of over 60 GFLOPs
at a power budget of 20 W.

A subsequent proposal by \cite{suda2016throughput} uses OpenCL to
implement
whole VGG \citep{Simonyan15} net on an Altera Stratix V board. Their throughput-optimized design
achieves an overall throughput of 117.8 GOPs. Finally a recent Xilinx-based
implementation \citep{Qiu:2016:GDE:2847263.2847265} achieves the start-of-art throughput of
137 GOPs. Their 16-bit fixed point implementation requires less than 10 W.

\subsection{Custom Accelerators (ASIC)}

Custom architectures have the highest throughput and energy efficiency, however their
design time is significant. DaDianNao by \cite{chen2014dadiannao} is a super-computer for
machine learning
at 28 nm technology. Their chip relies on large on-chip memory (which takes up nearly
half of the area) and achieves significant speedups and power savings compared to the
GPU. A later implementation termed Eyeriss \citep{isscc_2016_chen_eyeriss} can run the
convolutional layers of AlexNet in forward path at 34
frames per second (74.6 GOPs), using only 0.278~W. The chip is about 2X slower than a throughput
optimized embedded
GPU, but 13X more energy efficient. Eyeriss uses 16-bit fixed point. Finally EIE
\citep{han2016eie} is
an ASIC which leverages the deep compression pipeline \citep{han2015deep}. EIE infers
significantly
compressed networks 13X faster than a GeForce GTX Titan X. Since the neural network
is pruned and has shared weights, the whole network fits to on-chip memory which
allows to infer images using just 0.6 W.

\subsection{Comparison Of Accelerator Platforms}

In this section we compare different accelerator platforms in terms of throughput and throughput
per power. We take the performance numbers from recently published papers, the source of our
numbers can be found in Table \ref{tab:fpga_vs_gpu}.

\begin{table}[H]
\singlespacing
\caption[ASIC vs FPGA vs GPU]{Throughput and power consumption of different accelerator platforms.
All works implement an ImageNet network.}
\label{tab:fpga_vs_gpu}
\begin{center}
\begin{tabular}{*5l}
\toprule
\pbox{20cm}{ \bf{Platform} \\} & \pbox{20cm}{ \bf{Throughput} \\ }& \pbox{20cm}{ \bf{Power} \\ } & \pbox{20cm}{\bf{Throughput} \\ \bf{per power} } & \pbox{20cm}{ \bf{Source} \\ } \\
\midrule
ASIC & 74.6 GOP/s & 278 mW & 268 GOP/s/W & \cite{isscc_2016_chen_eyeriss} \\
\midrule
\pbox{20cm}{Xilinx Zynq ZC706} & 137 GOP/s & 9.63 W & 14.2 GOP/s/W & \cite{Qiu:2016:GDE:2847263.2847265} \\
\midrule
NVIDIA TK1 & 155 GOP/s & 10.2 W & 15.2 GOP/s/W & \cite{isscc_2016_chen_eyeriss} \\
\midrule
Titan X & 3.23 TOP/s & 250 W & 12.9 GOP/s/W & \cite{han2016eie} \\
\bottomrule
\end{tabular}
\end{center}
\end{table}

\begin{figure}[H]
\includegraphics[width=0.80\linewidth]{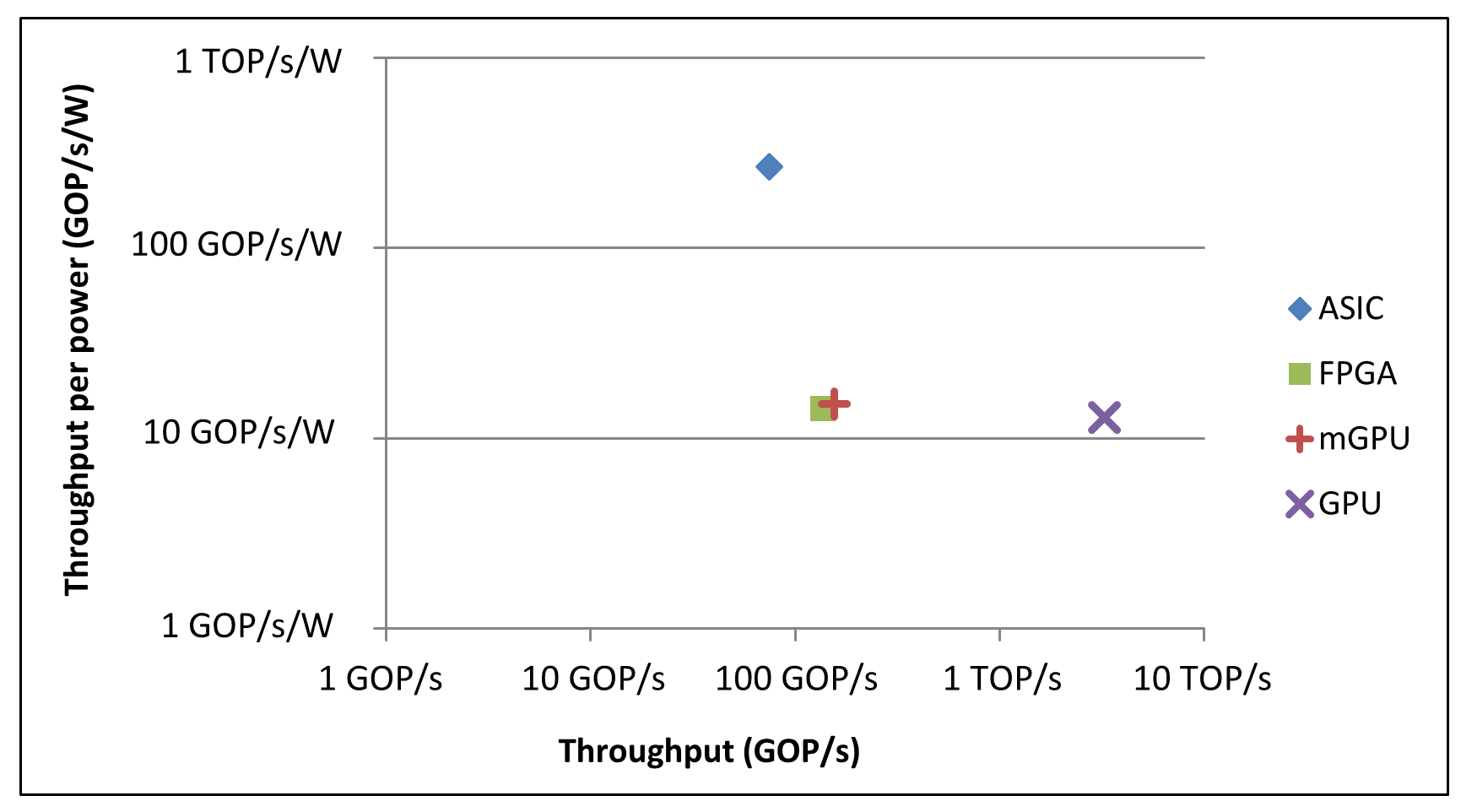}
\centering
\caption[ASIC vs FPGA vs GPU]
{Throughput and throughput per power for different accelerator platforms. Numbers are taken
from Table \ref{tab:fpga_vs_gpu}.}
\label{fig:fpga_vs_gpu}
\end{figure}

The ASIC design by \cite{isscc_2016_chen_eyeriss} is optimized for
large networks and low power consumption. Their work concentrates on convolutional layers of
AlexNet.
Other works which concentrate on fully connected layers only show similar throughput
\citep{han2016eie}. Predictably, the ASIC design shows the highest throughput per power (see
Figure \ref{fig:fpga_vs_gpu}).

For GPU performance, we consider an implementation which uses cuBLAS for fully connected layers of
AlexNet. Convolutional layers would yield lower throughput, since this layer type requires
rearrangement of data before the matrix-matrix multiplication. The acceleration on the GPU
achieves the highest throughput among all accelerators (see Figure \ref{fig:fpga_vs_gpu}). 
The high GPU throughput of 3.23 TOP/s comes
at a relatively high power cost (250 W). Embedded GPUs require less power, but their throughput
is proportionally lower. As a case in point, we consider the mobile GPU implementation of AlexNet
by \cite{isscc_2016_chen_eyeriss}. When comparing the two GPU implementations, the mobile
GPU's throughput per power is only slightly better than that of the high-end GPU
(15.2 GOP/s/W vs 12.9~GOP/s/W).

The FPGA implementation from \cite{Qiu:2016:GDE:2847263.2847265} is an end-to-end implementation
of the 16-layer version of VGG. The FPGA implementation uses 16-bit fixed point arithmetic
to reduce memory and computation requirements. Moreover the authors use pruning in
fully connected layers to reduce parameter size. Another work by \cite{suda2016throughput}
achieves nearly the same throughput without weight pruning. The FPGA implementation is head-to-head
with the embedded GPU implementation. The latter has 11\% more throughput and 6\% more throughput per
power.

\chapter{Fixed Point Approximation}
\label{chap:fixed_point}

This chapter covers the approximation of convolutional neural networks with fixed point numbers.
While normal inference is done in 32-bit floating point, using bit-width reduced format for
intermediate results can
increase both throughput and energy efficiency of hardware accelerators.

\section{Baseline Convolutional Neural Networks}

In the remainder of this document, we will discuss different approaches for
approximating CNNs in a hardware friendly manner. In each section, we approximate the
following CNNs:
\begin{enumerate}
  \item \textbf{LeNet}\footnote{\url{https://github.com/BVLC/caffe/blob/master/examples/mnist/lenet_train_test.prototxt}}
  was proposed by \cite{lecun1998gradient}. This network consists of two
convolutional and two fully connected layers and can be used to classify
handwritten digits (MNIST dataset).
  \item The CIFAR-10
data set \citep{krizhevsky2009learning} has 10 image classes such as airplanes, bird, and truck.
The \textbf{CIFAR-10 Full} model\footnote{\url{https://github.com/BVLC/caffe/blob/master/examples/cifar10/cifar10_full_train_test.prototxt}} 
was developed by Caffe for the CIFAR-10 data set. The network has three
convolutional layers followed by one fully connected layer. Moreover, the model
has two local response normalization (LRN) layers.
  \item \textbf{CaffeNet}\footnote{\url{https://github.com/BVLC/caffe/blob/master/models/bvlc_reference_caffenet/train_val.prototxt}}
is the Caffe version of AlexNet \citep{krizhevsky2012imagenet} which is the winner of the 2012
ILSVRC competition. This network can classify images into the 1000
ImageNet categories, which vary from animal and plant species to various human-made objects.
CaffeNet has five convolutional layers, three fully connected layers
and two LRN layers. CaffeNet has 60 million parameters and 650,000 neurons.
  \item \textbf{GoogLeNet}\footnote{\url{https://github.com/BVLC/caffe/wiki/Model-Zoo}} was proposed by
\cite{szegedy2015going} and won the 2014 ILSVRC
competition. This network is based on the inception idea, which uses
convolutional and pooling layers with small kernel sizes. GoogLeNet has 12X
fewer parameters than AlexNet but manages to improve the accuracy
significantly.
  \item \textbf{SqueezeNet}\footnote{\url{https://github.com/DeepScale/SqueezeNet/blob/master/SqueezeNet_v1.0/train_val.proto txt}}
by \cite{SqueezeNet} was developed with the goal of a small network with the accuracy of
AlexNet \citep{krizhevsky2012imagenet}. SqueezeNet relies on convolutional layers with 1x1 and
3x3
kernels. No fully connected layers or normalization layers are needed.
\end{enumerate}

\section{Fixed Point Format}

Standard inference of deep neural networks uses 32-bit floating point. We replace the
parameter and layer outputs with the following fixed point number format: [IL.FL],
where IL and FL denote the integer and fractional length of the numbers, respectively.
The number of bits used to represent each value is therefor IL+FL. To quantize floating
point numbers to fixed point, we use round-nearest. We use 2s-complement numbers,
thus the largest positive value we can represent is:

\begin{equation}
x_\mathit{max} = 2^{\mathit{IL}-1}-2^{-\mathit{FL}}
\end{equation}

Note that in the following experiments, all truncated numbers use a shared fixed point
format, i.e., they share the same integer and fractional length. For a representation using
dynamic adaption of integer and fractional part, please refer to chapter
\ref{chap:dyn_fixed_point}.

\section{Dynamic Range of Parameters and Layer Outputs}
In this subsection we analyze the dynamic range of numbers in two neural networks. This
analysis will help to understand the optimal choice for integer and fractional bits in fixed
point representation.

\subsection{Dynamic Range in Small CNN}

\begin{figure}[H]
\includegraphics[width=0.99\linewidth]{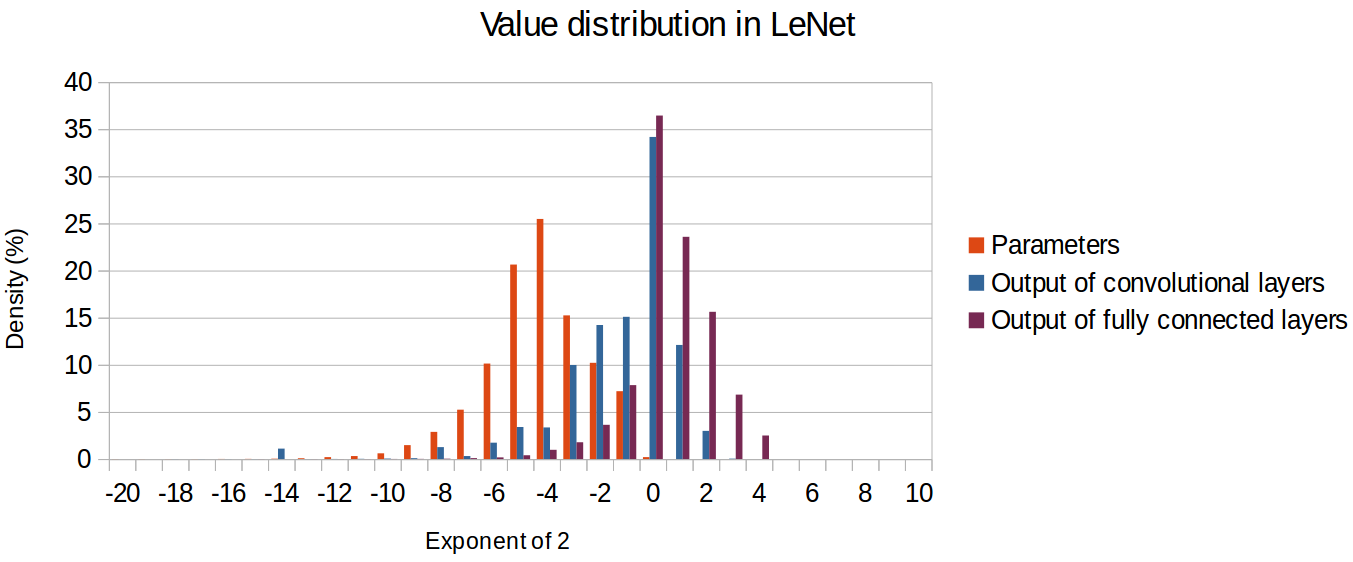}
\centering
\caption[Dynamic range of values in LeNet]{Dynamic range of values in LeNet.}
\label{fig:histograms_lenet}
\end{figure}

In a first step we do this analysis for LeNet. We performed the forward propagation of
100 images to compute intermediate values in the network. The value distribution is
shown in Figure \ref{fig:histograms_lenet}. Note that this histogram data is the result of
truncating all values
to integer power of two. We can see that on average, parameters are smaller than layer
outputs. 99\% of the trained network parameters are between $2^0$ and $2^{-10}$. For fully
connected layers however, 99\% of the layer outputs are in the range $2^5...2^{-4}$.

In order to quantize both the layer outputs and network parameters to 8-bit fixed point, a part
of the values needs to be
saturated. We achieved the best quantization results with the Q.4.4 format. This indicates
that large layer outputs are more important than small network parameters.

\subsection{Dynamic Range in Large CNN}
\label{chap:range_caffenet}

\begin{figure}[H]
\includegraphics[width=0.99\linewidth]{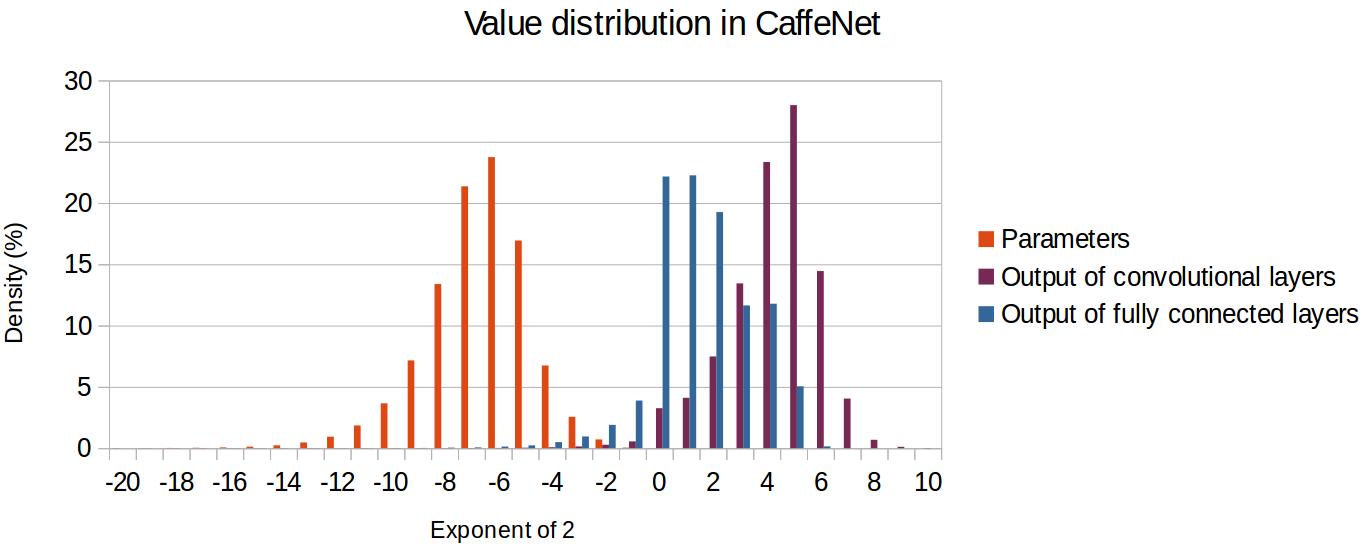}
\centering
\caption[Dynamic range of values in CaffeNet]{Dynamic range of values in CaffeNet.}
\label{fig:histograms_caffenet}
\end{figure}

This subsection contains the analysis for the relatively large CaffeNet network. We
performed the forward propagation of 50 images on a trained CaffeNet network. The resulting
dynamic range is shown in Figure \ref{fig:histograms_caffenet}.
Similarly to a small network, parameters tend to be smaller than layer outputs. However,
for this large network, the average difference between these two number categories is
much larger. This is to be expected, since layer outputs are the result of a multiplication
accumulation process, which yields a much larger result for big layers. As a case in point,
we can compare the relatively large second parameter layer (447.9M MAC operations) with the
relatively small last parameter layer (4.1M MAC operations). While the second
layer's largest value is larger than $2^9$, all values are below $2^6$ in the last layer.

Since the dynamic range of values is much larger than in LeNet, more bits are required
for a fixed point representations. Our experiments show the best 16-bit
fixed point results when using the Q9.7 format. Notice that a significant part of the network
parameters gets saturated in this quantization, since there are not enough fractional bits.
Only very few layer outputs (0.46\% in convolutional layers) are too large to be represented,
while a large part of the parameters (21.23\%) is truncated to zero. Similarly to the analysis with
LeNet, large layer outputs are more important than small parameters.

\section{Results}

This subsection covers the results of quantizing trained 32-bit floating point networks to fixed
point.

\subsection{Optimal Integer and Fractional Length}

For a given network and target bit-width, the layer outputs and network parameters of
convolutional and fully connected layers all share the same fixed point format. The bit-
width is the sum of integer and fractional length. The choice of fractional length is
crucial and will decide which values need to be saturated. Our quantization procedure
tries different partitionings of the bit-width into integer and fractional part. The best
setting is retained and the resulting fixed point network is fine-tuned. Notice that different
choices for integer and fractional length are conceivable. For example, only the layer
output quantization could be considered to find a good partitioning, since the network
parameters can be adapted in the fine-tuning step. However, our experiments on three
different networks show that a joint optimization of layer outputs and parameters yields
the best results after fine-tuning.

\subsection{Quantization to Fixed Point}

\begin{figure}[H]
\includegraphics[width=0.9\linewidth]{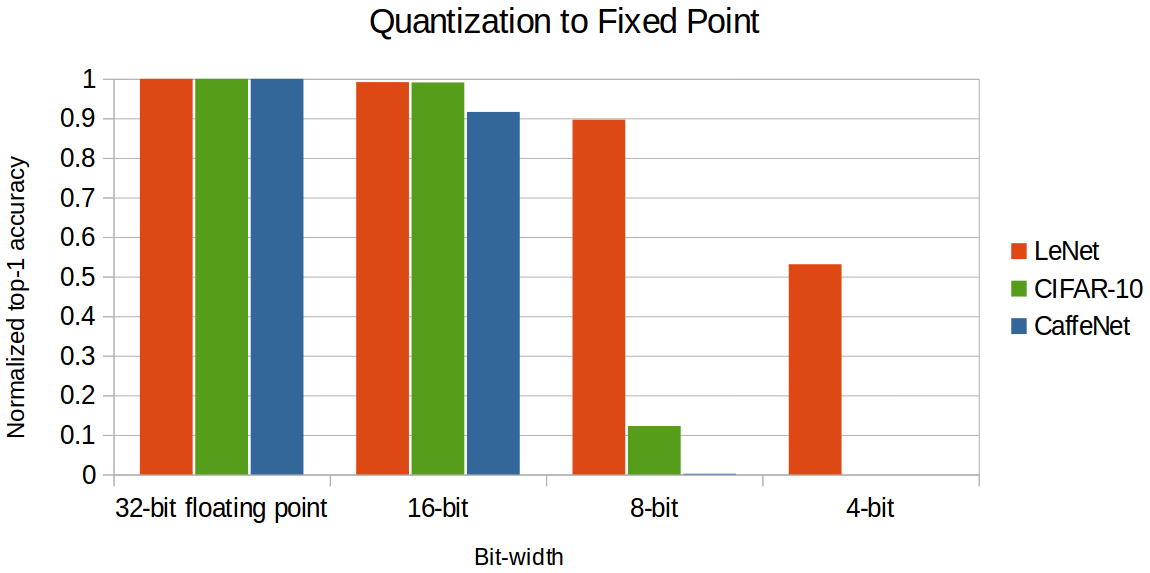}
\centering
\caption[Fixed point results]
{Normalized classification accuracy of fixed point networks.}
\label{fig:static_fixed_point}
\end{figure}

We quantized three of our baseline networks to fixed point: LeNet, CIFAR-10 and
CaffeNet.
To calculate relative accuracy of a bit-width reduced network, we divide the fixed point
accuracy
by the 32-bit floating point accuracy.
First we consider the relatively small LeNet network for handwritten digit
recognition.
The quantization from 32-bit floating point to 8-bit fixed point incurs a
relative accuracy loss of 10.3\% (see Figure \ref{fig:static_fixed_point}).
After fine-tuning, the absolute
accuracy loss shrinks to 0.27\% (Table \ref{tab:fixed_point_res}), indicating LeNet works well
in 8-bit fixed point.

The second baseline network we consider is CIFAR-10. This network classifies images
into classes such as truck, ship, dog, bird. As this is a more challenging task which
requires a larger network, the layer outputs are larger too, and the network is more
sensitive to quantization errors. When the network is quantized to 8-bit, the network
output is random and the accuracy drops to 10\%. Since our quantization framework is
unable to achieve good results at 8-bit, we double the bit-width. The network works fine in
16-bit, with a relative accuracy loss below 1\%. The best results were achieved using 8
integer bits, whereas LeNet only required 4 integer bits (see Table \ref{tab:fixed_point_res}).

\begin{table}[H]
\singlespacing
\caption[Fixed point results]{Fine-tuned fixed point networks.
Numbers in brackets indicate accuracy without fine-tuning.}
\label{tab:fixed_point_res}
\begin{center}
\begin{tabular}{*5l}
\toprule
\pbox{20cm}{\bf{Network} \\} & \pbox{20cm}{\bf{Baseline} \\ \bf{accuracy}} & \pbox{20cm}{\bf{Fixed point}\\ \bf{bit-width}} & \pbox{20cm}{\bf{Fixed point} \\ \bf{format}} & \pbox{20cm}{\bf{Fixed point} \\ \bf{accuracy}} \\
\midrule
LeNet & 99.15\% & 8-bit & Q4.4 & 98.88\% (88.90\%) \\
CIFAR-10 & 81.69\% & 16-bit & Q8.8 & 81.38\% (80.94\%) \\
CaffeNet top-1 & 56.90\% & 16-bit & Q9.7 & 52.48\% (52.13\%) \\
\bottomrule
\end{tabular}
\end{center}
\end{table}

Finally we quantize CaffeNet, a network for ImageNet classification. As before, 8-bit
quantization yields poor results, which is the reason we choose 16-bit fixed point.
The relative accuracy loss after quantization is 8.4\% and the fine-tuned network achieves
an accuracy within 4.5\% of the baseline (compared in absolute values).

In order to increase the accuracy of the quantized networks, we introduce dynamic fixed point in
the next section.
\chapter{Dynamic Fixed Point Approximation}
\label{chap:dyn_fixed_point}

In this chapter we discuss quantization of a floating point CNN to a dynamic
fixed point version. We extend the fixed point format to dynamic fixed point, and show
how it can be used to further decrease parameter size while maintaining a high prediction
accuracy.

\section{Mixed Precision Fixed Point}

\begin{figure}[H]
\includegraphics[width=0.8\linewidth]{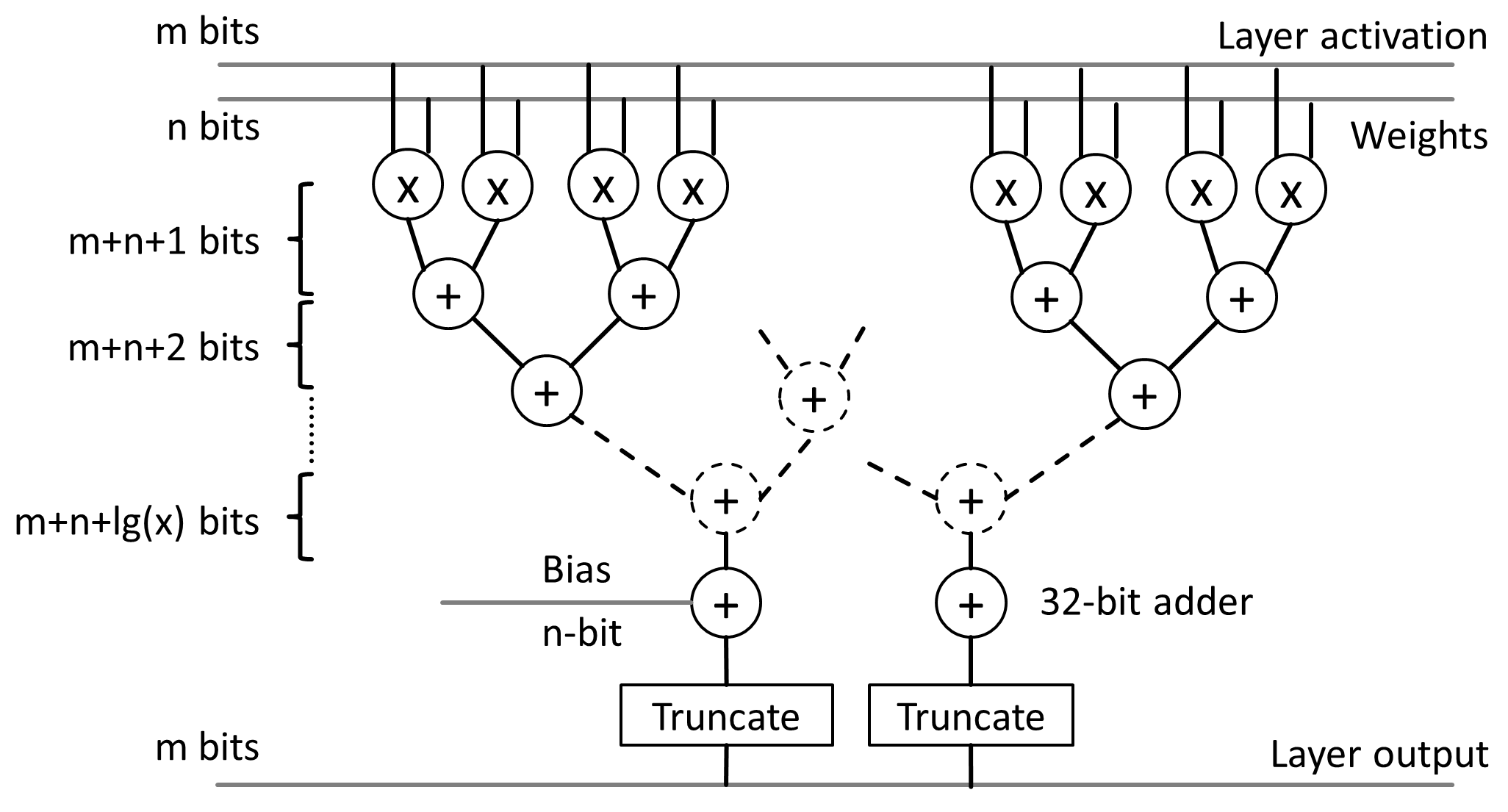}
\centering
\caption[Fixed point data path]{Data path of a fixed point convolutional or fully connected layer.}
\label{fig:data_path_dynamic_fixed_point}
\end{figure}

The data path of fully connected and convolutional layers consists of a series of MAC
operations (multiplication and accumulation), as shown in Figure
\ref{fig:data_path_dynamic_fixed_point}. The layer
activations are multiplied with the network weights, and these multiplication results are
accumulated to form the output.

As shown by \cite{lin2015fixed, Qiu:2016:GDE:2847263.2847265}, it is a good approach to use mixed
precision, i.e., different
parts of a CNN use different bit-widths. In Figure \ref{fig:data_path_dynamic_fixed_point},
$m$ and $n$ refer to the number of bits used to represent
layer outputs and layer weights, respectively. Multiplication results are accumulated
using an adder tree which gets thicker towards the end. The adder outputs in the first level
are $m + n + 1$ bits wide, and the bit-width grows by 1 bit in each level. In the last level,
the bit-width is $m + n + \lg_2 (x)$, where $x$ is the number of multiplication operations per
output value. In the last stage, the bias is added to form the layer output. For each
network layer, we need to find a good balance between reducing the bit-widths ($m$ and $n$)
and maintaining a good classification accuracy.

\section{Dynamic Fixed Point}

The different parts of a CNN have a significant dynamic range. In large layers, the
outputs are the result of thousands of accumulations, thus the network parameters are
much smaller than the layer outputs. Fixed point has only limited capability to cover a
wide dynamic range. Dynamic fixed point can be a good solution to overcome this
problem, as shown by \cite{courbariaux2014low}. In dynamic fixed point, each
number is represented as follows:

\begin{equation}
(-1)^s \cdot 2^{-FL} \sum_{i=0}^{B-2} 2^i \cdot x_i
\end{equation}

Here $B$ denotes the bit-width, $s$ the sign bit, $FL$ is the fractional length, and $x$ the
mantissa
bits. Since the intermediate values in a network have different ranges, it is desirable to
\textit{group fixed point numbers into groups with constant FL}. So the number of bits allocated to
the fractional part is constant within that group, but different compared to other groups.
Each network layer is split into two groups: one for the layer outputs, one for the layer
weights. This allows to better cover the dynamic range of both layer outputs and weights,
as weights are normally significantly smaller. On the hardware side, it is possible to
realize dynamic fixed point arithmetic using bit shifters.

\begin{figure}[H]
\includegraphics[width=0.4\linewidth]{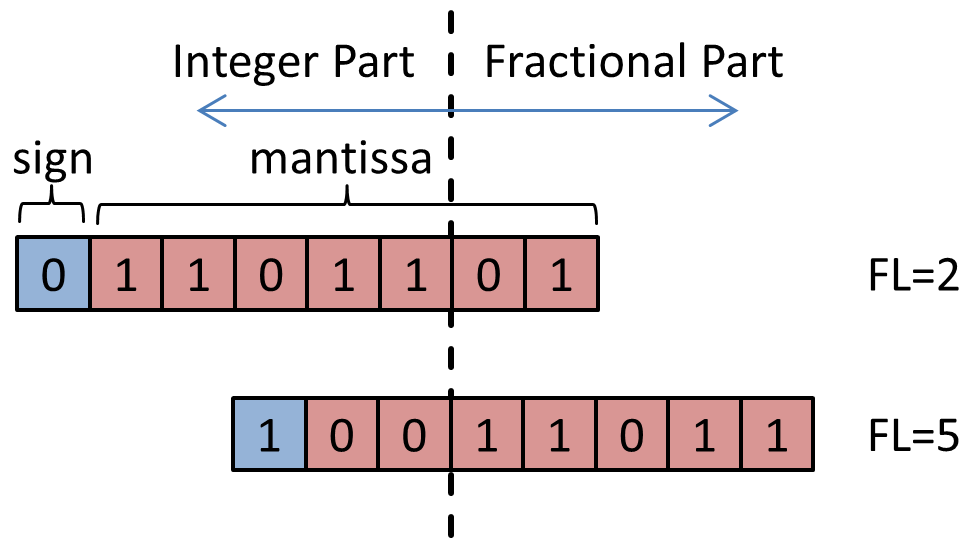}
\centering
\caption[Dynamic fixed point representation]{Dynamic fixed point with variable length of
fractional part.}
\label{fig:dynamic_fixed_point}
\end{figure}

The concept of dynamic fixed point is depicted in Figure \ref{fig:dynamic_fixed_point},
where two numbers are
both represented in 8 bits, but belong to a different group (i.e., they have different
fractional length).

\subsection{Choice of Number Format}

When we approximate a neural network with dynamic fixed point numbers, we need to
choose a number format for each number group. Each layer has two such groups: the
layer parameters and the layer outputs. Within each group, all numbers are
represented using the same integer and fractional length.

To find the optimal set of number formats, we could perform an exhaustive search,
however this is not efficient for large neural networks. Instead, we follow a specific rule
that automatically determines the required number of integer bits. More specifically, we
choose enough bits to avoid saturation. So for a given set of numbers $S$, the required
integer length \textit{IL} is given by Equation \ref{eq:il_dyn_fixed_point}.

\begin{equation}
\label{eq:il_dyn_fixed_point}
\mathit{IL} = \lceil \lg_2(\max_{S}x+1) \rceil
\end{equation}

This relation defines the integer length of layer parameters. For layer outputs, we reduce the
integer length by one, since our experiments
show slightly better results this way.

\section{Results}

In this section we present the results of approximating 32-bit floating point networks by
condensed dynamic fixed point models. All classification accuracies were obtained running the
respective network on the whole validation dataset. We follow the general approximation
procedure explained in section \ref{chap:nn_limited_precision}.

\subsection{Impact of Dynamic Fixed Point}

\begin{figure}[H]
\includegraphics[width=0.8\linewidth]{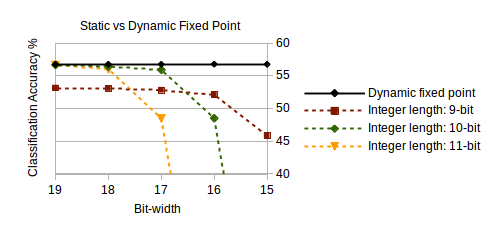}
\centering
\caption[Static vs dynamic fixed point]{Top-1 accuracy for CaffeNet on ILSVRC 2014 validation
dataset. Integer
length refers to the number of bits assigned to the integer part of fixed point numbers.}
\label{fig:static_vs_dynamic2}
\end{figure}

We used our Ristretto framework to quantize CaffeNet (AlexNet) into fixed point, and
compare traditional fixed point with dynamic fixed point. To allow a simpler comparison,
all layer outputs and network parameters share the same bit-width. Results show a good
performance of static fixed point for as low as 18-bit (Figure \ref{fig:static_vs_dynamic2}).
However, when
reducing the bit-width further, the accuracy starts to drop significantly, while dynamic
fixed point has a stable accuracy. We can conclude that dynamic fixed point performs
significantly better for such a large network. The reason is that dynamic fixed point
allows us to adapt the number of bits allocated to integer and fractional part, according to
the dynamic range of different parts of the network.

\subsection{Quantization of Individual Network Parts}

In this section, we present the results for approximating different parts of a network.
For each experiment, only one category is quantized to dynamic fixed point, and the rest remains
in full precision.
Table~\ref{tab:dyn_fixed_point_quantization} shows the quantization impact for three different
networks. For each network, we quantize the layer outputs, the convolutional kernels (CONV),
and the parameters of fully connected layers (FC) independently.
In all three nets, the convolution kernels and layer activations can be trimmed to 8-bit
with an absolute accuracy change of only 0.3\%. Fully connected layers are more affected
from trimming weights to 8-bit, the absolute change is maximally 0.9\%. Interestingly,
LeNet weights can be trimmed to as low as 2 bits, with absolute accuracy change below
0.4\%.

\begin{table}[H]
\caption[Dynamic fixed point quantization]{Dynamic fixed point quantization results for
different parts of network. Only one
number category is cast to fixed point, and the remaining numbers are in floating point
format.}
\singlespacing
\label{tab:dyn_fixed_point_quantization}
\begin{center}
\begin{tabular}{*5l}
\toprule
Fixed point bit-width & 16-bit & 8-bit & 4-bit & 2-bit \\
\midrule
\multicolumn{5}{l}{\textbf{LeNet, 32-bit floating point accuracy: 99.1\%}} \\
\midrule
Layer output & 99.1\% & 99.1\% & 98.9\% & 85.9\%\\
CONV parameters & 99.1\% & 99.1\% & 99.1\% & 98.9\%\\
FC parameters & 99.1\% & 99.1\% & 98.9\% & 98.7\%\\
\midrule
\multicolumn{5}{l}{\textbf{Full CIFAR-10, 32-bit floating point accuracy: 81.7\%}} \\
\midrule
Layer output & 81.6\% & 81.6\% & 79.6\% & 48.0\%\\
CONV parameters & 81.7\% & 81.4\% & 75.9\% & 19.1\%\\
FC parameters & 81.7\% & 80.8\% & 79.9\% & 77.5\%\\
\midrule
\multicolumn{5}{l}{\textbf{CaffeNet Top-1, 32-bit floating point accuracy: 56.9\%}} \\
\midrule
Layer output & 56.8\% & 56.7\% & 06.0\% & 00.1\% \\
CONV parameters & 56.9\% & 56.7\% & 00.1\% & 00.1\%\\
FC parameters & 56.9\% & 56.3\% & 00.1\% & 00.1\%\\
\bottomrule
\end{tabular}
\end{center}
\end{table}

\subsection{Fine-tuned Dynamic Fixed Point Networks}
\label{chap:dyn_fixed_point_res}

Here we report the accuracy of five networks that were condensed and fine-tuned with
Ristretto. All networks use dynamic fixed point parameters as well as dynamic fixed
point layer outputs for convolutional and fully connected layers. LeNet performs well in
2/4-bit, while CIFAR-10 and the three ImageNet CNNs can be trimmed to 8-bit (see
Table \ref{tab:dyn_fixed_point_finetune}). Surprisingly, these compressed networks still perform
nearly as well as their
floating point baseline. The relative accuracy drops of LeNet, CIFAR-10 and SqueezeNet
are very small ($<$0.6\%), whereas the approximation of the larger CaffeNet and
GoogLeNet incurs a slightly higher cost (0.9\% and 2.3\% respectively).

\begin{table}[H]
\caption[Dynamic fixed point results]{Fine-tuned, dynamic fixed point networks. Numbers
in brackets indicate
accuracy without fine-tuning.}
\singlespacing
\label{tab:dyn_fixed_point_finetune}
\begin{center}
\begin{tabular}{*6l}
\toprule
\pbox{20cm}{\bf{Network}\\} & \pbox{20cm}{\bf{Layer} \\ \bf{outputs}} & \pbox{20cm}{\bf{CONV} \\ \bf{parameters}} & \pbox{20cm}{\bf{FC}\\ \bf{parameters}} & \pbox{20cm}{\bf{32-bit} \\ \bf{baseline}} & \pbox{20cm}{\bf{Fixed point} \\ \bf{accuracy}} \\
\midrule
LeNet (Exp 1) & 4-bit & 4-bit & 4-bit & 99.15\% & 98.95\% (98.72\%) \\
LeNet (Exp 2) & 4-bit & 2-bit & 2-bit & 99.15\% & 98.81\% (98.03\%) \\
Full CIFAR-10 & 8-bit & 8-bit & 8-bit & 81.69\% & 81.44\% (80.64\%) \\
CaffeNet & 8-bit & 8-bit & 8-bit & 56.90\% & 56.00\% (55.77\%) \\
SqueezeNet & 8-bit & 8-bit & 8-bit & 57.68\% & 57.09\% (55.25\%) \\
GoogLeNet & 8-bit & 8-bit & 8-bit & 68.92\% & 66.57\% (66.07\%) \\
\bottomrule
\end{tabular}
\end{center}
\end{table}

The SqueezeNet \citep{SqueezeNet} architecture was developed with the goal of a small CNN that
performs well on the ImageNet data set. Ristretto can make the already small network
even smaller, so that its parameter size is less than 2 MB. This condensed network is well-
suited for deployment in smart mobile systems.

All five 32-bit floating point networks can be approximated well in 8-bit and 4-bit fixed
point. For a hardware implementation, this reduces the size of multiplication units by
about one order of magnitude. Moreover, the required memory bandwidth is reduced by
4--8X. Finally, it helps to hold 4--8X more parameters in on-chip buffers.

Some previous work \citep{courbariaux2014low} concentrated on training with fixed point arithmetic from the start
and shows little performance decline for as short as 7-bit fixed
point numbers on
LeNet. Our approach is different in that we train with high numerical precision, then
quantize to fixed point, and finally fine-tune the fixed point network. Our condensed
model achieves superior accuracy with as low as 4-bit fixed point, on the same data set.
While more sophisticated data compression schemes could be used to achieve higher
network size reduction, our approach is very hardware friendly and imposes no additional
overhead such as decompression.

\chapter{Minifloat Approximation}

\section{Motivation}

Chapters \ref{chap:fixed_point} and \ref{chap:dyn_fixed_point} concentrated on fixed point
approximation of deep CNNs. Since the
training of neural networks is normally done in floating point, it is an intuitive approach
to condense these models to smaller floating point numbers. This section analyses the
network approximation through minifloat, i.e., floating point numbers with 16
bits or smaller.

\section{IEEE-754 Single Precision Standard}

According to IEEE-754 standard, single precision numbers have 1 sign bit, 8 exponent
bits and 23 mantissa bits. The mantissa's first bit (always '1') is added implicitly, and the
stored exponent is biased by 127. Numbers with all zeros or ones in the exponent have a
special meaning. An exponent with all zeros either represents the number 0 or a
denormalized number, depending on the mantissa bits. For the case of all ones in the
exponent, the number represented is either +/-INF or NaN.

\section{Minifloat Number Format}

In order to condense networks and reduce their computational and memory requirements,
we will represent floating point numbers with much fewer bits than the IEEE-754
standard. We follow the standard to a large degree when going to 12-bit, 8-bit, or 6-bit
numbers, but our format differs in some details. Namely, the exponent bias
is lowered according to the number of bits assigned to the exponent:

\begin{equation}
\mathit{bias} = 2^{\mathit{bits}-1}-1
\end{equation}

\begin{figure}[H]
\includegraphics[width=0.5\linewidth]{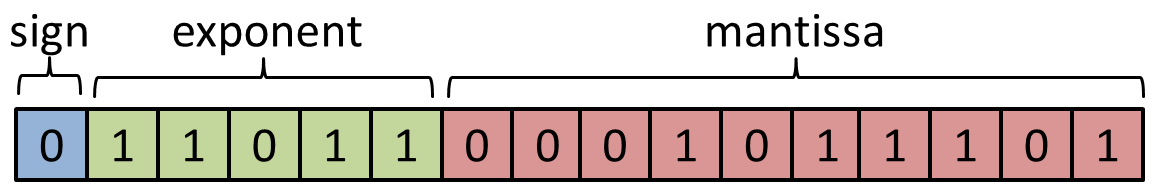}
\centering
\caption[Minifloat number representation]{Example of a 16-bit minifloat
number with 5 exponent bits and 10 mantissa bits.}
\label{fig:mini_floating_point}
\end{figure}

Here $bits$ denotes the number of bits assigned to the exponent. Another difference to
the IEEE standard is that we don't support denormalized numbers, INF and NaN. INF is
replaced by saturated numbers, and denormalized numbers are replace by 0. Finally, the
number of bits assigned to the exponent and mantissa part don't follow a specific rule. To
be more precise, our Ristretto framework automatically searches for the best balance
between exponent and mantissa bits. As a case in point, a 16-bit minifloat
number (Figure \ref{fig:mini_floating_point}) could be represented with 1 sign bit,
5 exponent bits and 10
mantissa bits.

\subsection{Network-specific Choice of Number Format}

Similar to dynamic fixed point, we need to choose a specific number format per bit-width,
i.e., partition the available bits into exponent and mantissa. In order to approximate
a neural network with minifloat numbers, we need to
find a suitable number of exponent and mantissa bits. We use enough exponent $bits$ to
avoid saturation:

\begin{equation}
\mathit{bits} = \lceil \lg_2(\lg_2(\max_S x)-1)+1 \rceil
\label{eq:minifloat_exp_bits}
\end{equation}

$S$ is the set of numbers which we approximate. This choice of exponent bits assures no
saturation happens, under the assumption that we use infinitely many mantissa bits.

\section{Data Path for Accelerator}

\begin{figure}[H]
\includegraphics[width=0.7\linewidth]{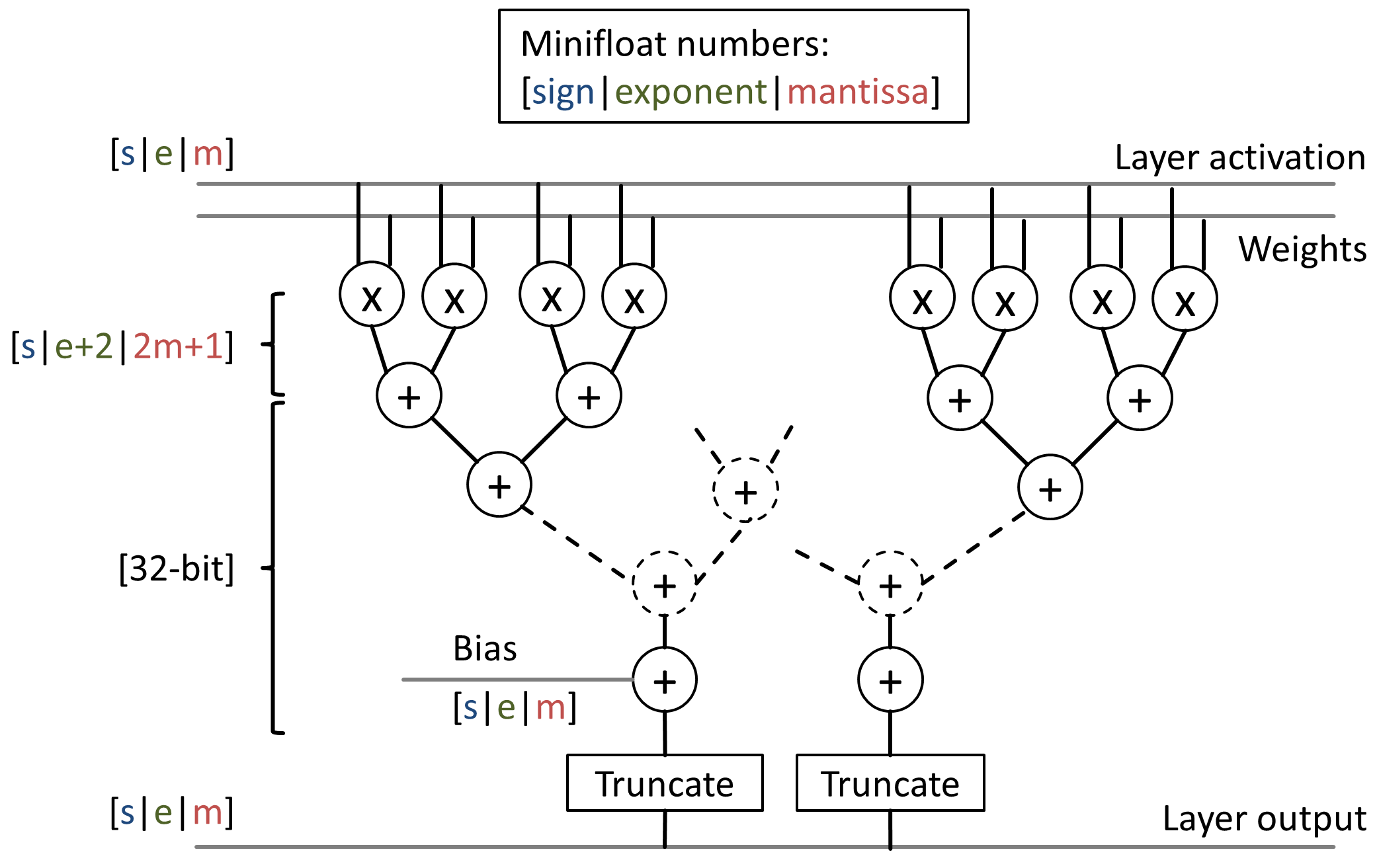}
\centering
\caption[Minifloat data path]{Data path of a minifloat convolutional or
fully connected layer.}
\label{fig:data_flow_fp}
\end{figure}

The data path of convolutional and fully connected layers is depicted in Figure
\ref{fig:data_flow_fp}. For
simplicity, we only consider fixed precision arithmetic, i.e., all number categories shared
the same minifloat format. Similar to the fixed point data path, network
parameters and layer inputs are multiplied and accumulated. Input to each multiplier is a
pair of numbers, each in minifloat format. The output of each multiplier is 3 bits
wider than the input numbers. In a next step, the multiplication results are accumulated in
full precision. In a last step the bias is added in minifloat format, and the final
result is trimmed to minifloat.

When implemented in a hardware accelerator, the data path's input and output are minifloat
numbers. In case the neural network in question is too large to fit into on-chip
memory, the layer outputs and parameters need to be stored in off-chip memory.
Since both these number categories are represented in minifloat, we can achieve
significant energy savings thanks to reduced data transfer.

\section{Results}

In this section, we analyze the impact of lowering the bit-width of floating point numbers.
We used our approximation framework to query different CNNs which use minifloat numbers in
convolutional and fully connected layers. To find the accuracy
of the condensed networks, we follow the quantization flow described in section
\ref{chap:nn_limited_precision}.

\begin{figure}[H]
\includegraphics[width=0.8\linewidth]{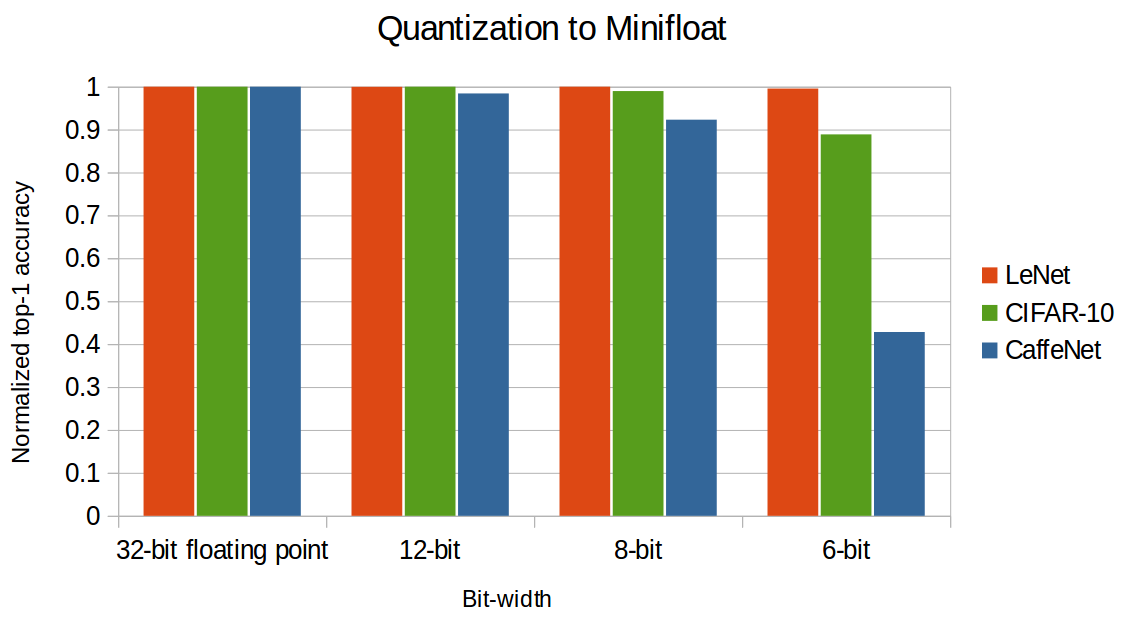}
\centering
\caption[Minifloat results]{Normalized classification accuracy of minifloat
networks.}
\label{fig:mini_fp.png}
\end{figure}

We quantized three CNNs to 12, 8 and 6-bit minifloat. The quantization is done
for layer outputs and parameters of fully connected and convolutional layers. For each
network, we show the classification accuracy of both the 32-bit baseline, followed by
minifloat versions (Figure \ref{fig:mini_fp.png}). We calculate the normalized
accuracy by dividing the minifloat network's performance by the 32-bit floating point network
accuracy.

The results indicate that LeNet has no classification loss when shrinking layer outputs
and layer parameters to 8-bit. CIFAR-10 and CaffeNet can be used in 12-bit, again with
no loss in accuracy.

We fine-tuned the 8-bit versions of the three networks. Table 
\ref{tab:mini_fp_res} shows
the accuracy of the minifloat networks. CIFAR-10 has an absolute accuracy
drop below 1\%, and CaffeNet incurs an absolute drop of 4.6\%. For LeNet, minifloat
actually increases accuracy. Minifloat adds more regularization to LeNet
and increases the accuracy by 0.05\%, compared to the 32-bit network.

\begin{table}[H]
\singlespacing
\caption[Minifloat results]{Fine-tuned minifloat networks. Numbers in
brackets indicate accuracy
without fine-tuning.}
\label{tab:mini_fp_res}
\begin{center}
\begin{tabular}{*5l}
\toprule
\pbox{20cm}{\bf{Network} \\} & \pbox{20cm}{\bf{32-bit} \\ \bf{accuracy}} & \pbox{20cm}{\bf{Minifloat}\\ \bf{bit-width}} & \pbox{20cm}{\bf{Minifloat} \\ \bf{accuracy}} & \pbox{20cm}{\bf{Exponent bits,}\\ \bf{mantissa bits}} \\
\midrule
LeNet & 99.15\% & 8-bit & 99.20\% (99.20\%) & 4-bit, 3-bit \\
CIFAR-10 & 81.69\% & 8-bit & 80.85\% (80.47\%) & 5-bit, 2-bit \\
CaffeNet top-1 & 56.90\% & 8-bit & 52.52\% (52.30\%) & 5-bit, 2-bit \\
\bottomrule
\end{tabular}
\end{center}
\end{table}

\newpage
Compared to dynamic fixed point results in section \ref{chap:dyn_fixed_point_res}, minifloat
requires
more bits. For 8-bit CaffeNet, the absolute accuracy drop of dynamic fixed point is small
(below 1\%), whereas minifloat incurs a relatively large drop (4.38\%).

\section{Comparison to Previous Work}
Previous work \citep{deng2015reduced} approximated network parameters of AlexNet (CaffeNet)
with 8-bit
minifloat. They analyze the impact of quantizing a varying percentage of the
network parameters. Our results achieve significantly better accuracy, thanks to a careful
choice of minifloat format and a fine-tuning step.

\chapter{Turning Multiplications Into Bit Shifts}

\section{Multiplier-free Arithmetic}

Hardware accelerators for convolutional neural networks need to be energy-efficient to
allow for deployment in mobile devices. Fully connected layers and convolutional layers
consist of additions and multiplications, of which the latter requires a much larger chip
area. This motivated previous research to eliminate all multipliers by using \textit{integer power
of two} weights \citep{tang1993multilayer, mahoney2008backpropagation}. These weights can be
considered as minifloat numbers with zero mantissa bits. By using such weights, all multiplications
turn into bit shifts, which can save a significant amount of energy on a hardware accelerator.

We now detail the approximation of convolutional and fully connected layers with multiplier-free
arithmetic. Although we assume strictly positive weights in this discussion, it is straight
forward
the expand the approximation procedure to both positive and negative weights.
The computation of convolutional and fully connected layers consists of
multiplication-and-accumulation operations. Equation \ref{eq:layer_with_mults} shows the necessary
operations in a full precision network. The layer inputs $x_j$ are multiplied with layer
parameters $w_j$
and the accumulation yields the result $z_i$. To simplify this discussion, we assume the input
data has been rearranged such that output $z_i=\textbf{w}^T\cdot \textbf{x}$.
In order to switch to multiplier-free arithmetic, we first
approximate parameters by the closest integer-power-of-two
number (Equation \ref{eq:to_power_2}). Now output $z_i$ can be approximated by equation 
\ref{eq:layer_no_mults} which is multiplier-free.
Notice that the last equation relies on the
power-of-two exponents $e_j$, not the original parameters.

\begin{equation}
 \label{eq:layer_with_mults}
 z_i=\sum_j x_j \cdot w_j
\end{equation}

\begin{equation}
 \label{eq:to_power_2}
 e_j=round(\lg_2(w_j))
\end{equation}

\begin{equation}
 \label{eq:layer_no_mults}
 z_i \approx \sum_j x_j \texttt{<<} e_j
\end{equation}

\section{Maximal Number of Shifts}

Nearly all network weights of a trained network are between $+1$ and $-1$, but most of them
are close to zero. The quantization of these parameters to power-of-two has the highest
impact (in terms of absolute value change) to the weights close to $+1$ and $-1$. These
weights will only be able to take on the values $1$, $\frac{1}{2}$, $\frac{1}{4}$ and so on.

\begin{figure}[H]
\includegraphics[width=0.15\linewidth]{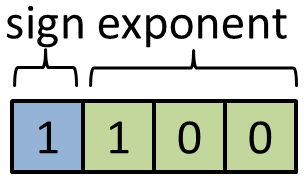}
\centering
\caption[Representation for integer-power-of-two parameter]{Example of a 4-bit parameter.}
\label{fig:power-2-weight}
\end{figure}

We encode the number of shifts in 4 bits (see Figure \ref{fig:power-2-weight}). This implies
the parameter exponents can have 8 different values.
We choose to represent the exponent values such that $e_i \in [-8,...,-1]$ and use this format
for the subsequent
experiments.

The motivation for this format is two-fold. First of all, using only 4 bits for parameters
reduces the memory requirements tremendously. Second, the smallest possible value in
this format is $2^{-8}$. Parameters smaller than that have only a minor effect on the network
output. Moreover only few parameters are below this smallest value. Our analysis from
section \ref{chap:range_caffenet} shows only 10.97\% of parameters in CaffeNet are lower
than the smallest possible value. For
LeNet, this percentage is even smaller (5.83\%).

\section{Data Path for Accelerator}

\begin{figure}[H]
\includegraphics[width=0.7\linewidth]{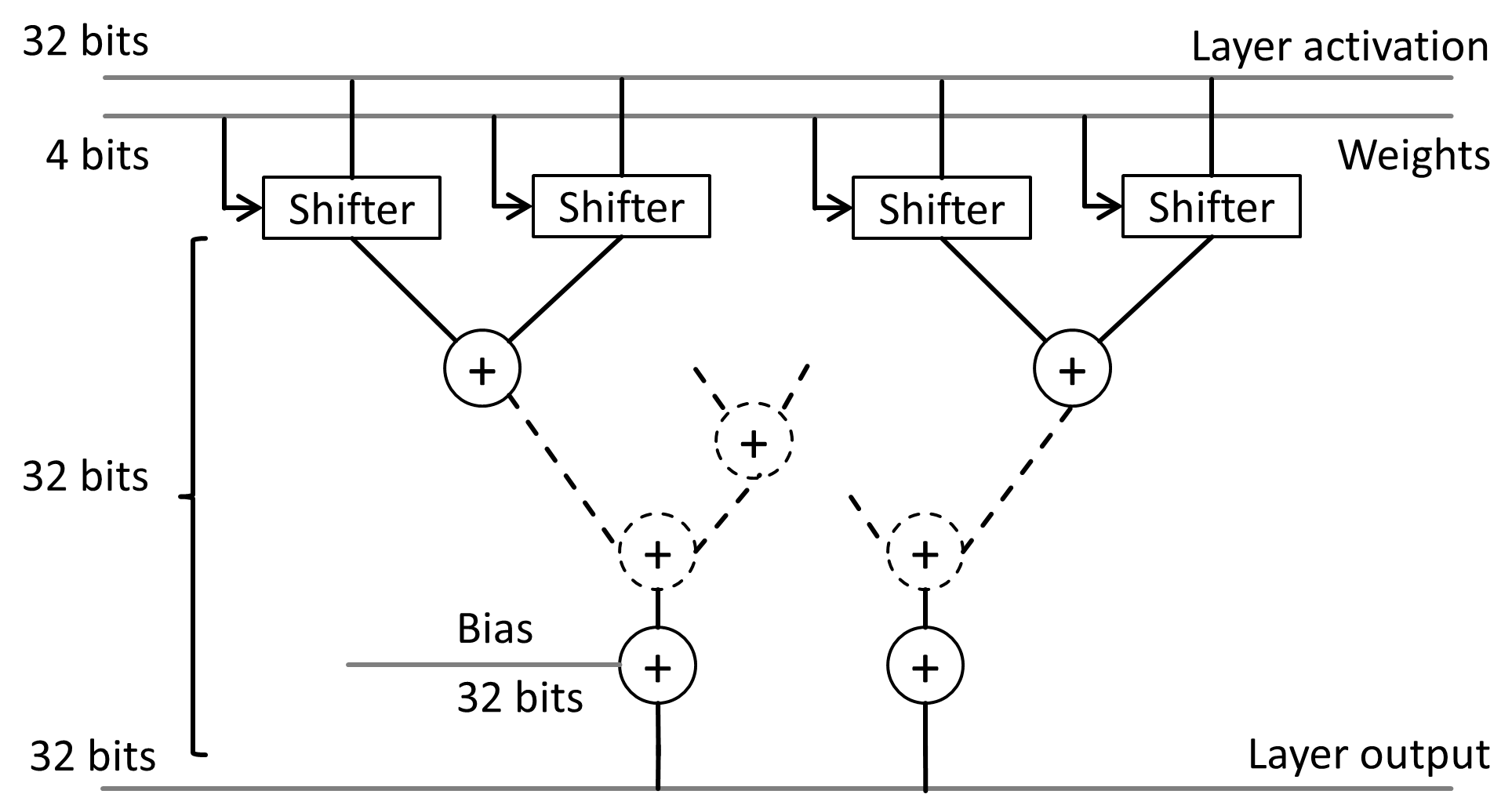}
\centering
\caption[Multiplier-free data path]{Data path of a convolutional or fully
connected layer.}
\label{fig:power_2_data_path}
\end{figure}

The adder-only arithmetic of convolutional and fully connected layers is shown
Figure \ref{fig:power_2_data_path}. The 4-bit parameters indicate how many bit-shifts are
required for the
layer inputs. To enable shifts by multiple bits in one clock cycle, barrel shifters should be
used. Notice that this data path has no multipliers at all, which can potentially save
significant chip area. To simplify this analysis, we only focus on the impact of removing
multiplications. The layer inputs and outputs are kept in full precision format.

\section{Results}

\begin{table}[H]
\singlespacing
\caption[Multiplier-free arithmetic results]
{Classification accuracy of nets with power-of-two weights. Numbers in brackets
indicate accuracy without fine-tuning.}
\label{tab:power_2_res}
\begin{center}
\begin{tabular}{*3l}
\toprule
\pbox{20cm}{\bf{Network} \\} & \pbox{20cm}{\bf{32-bit floating} \\ \bf{point weights}} & \pbox{20cm}{\bf{Multiplier-free}\\ \bf{network}} \\
\midrule
LeNet & 99.15\% & 99.16\% (99.09\%) \\
CIFAR-10 & 81.69\% & 77.48\% (74.65\%) \\
CaffeNet top-1 & 56.90\% & 53.25\% (50.58\%) \\
\bottomrule
\end{tabular}
\end{center}
\end{table}

We used our Ristretto framework to simulate the effect of removing all multiplications from
convolutional and fully connected layers. Our framework quantizes all network parameters to the
nearest integer-power-of-two number. Table \ref{tab:power_2_res} compares networks with
power-of-two weights and networks with single precision weights.

For LeNet, the absolute classification accuracy drop for the quantized
weights is 0.1\%. CIFAR-10 and CaffeNet are more affected by the weight
quantization (4.21\% and 3.65\% absolute accuracy drop). At first glance, the results
for the larger two networks might be
discouraging. However, it is surprising that the nets with weight quantization still have a
decent classification accuracy. The power-of-two weights can be stored in just 4 bits (the
exponents range from -1 to -8). This allows for tremendous energy savings: First the
traffic to off-chip memory is reduced, as the weights are not 32-bit but 4-bit. Second,
multipliers are replaced with simple bit shifters.
\chapter{Comparison of Different Approximations}

\begin{figure}[H]
\includegraphics[width=0.9\linewidth]{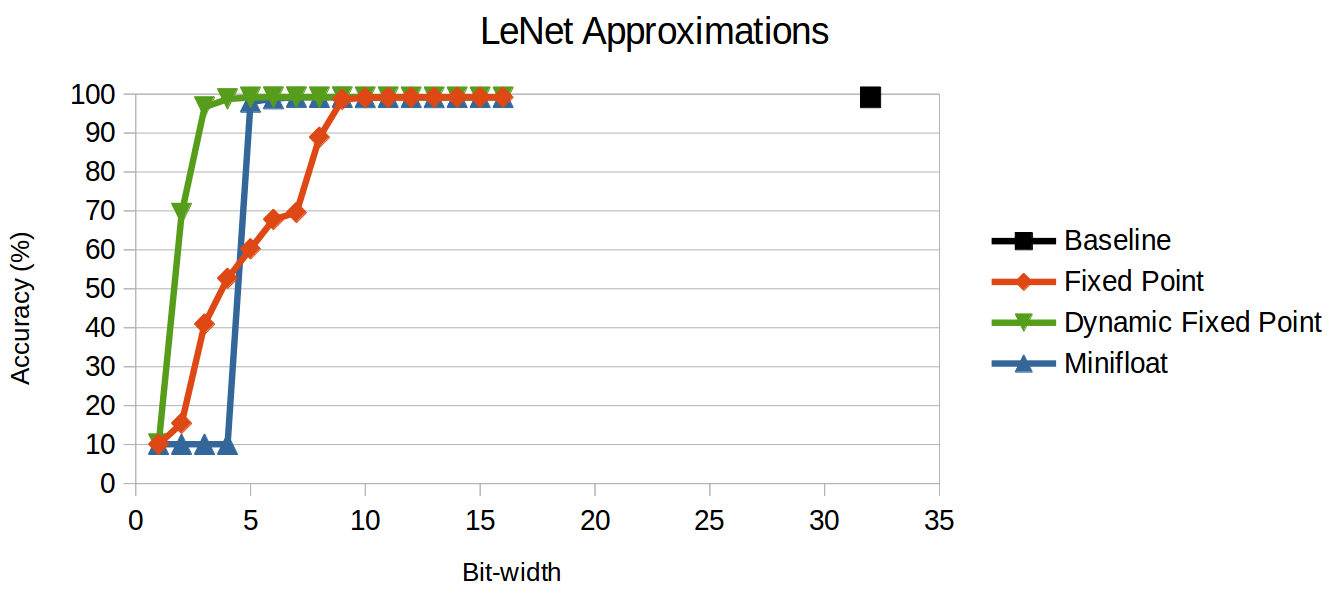}
\centering
\caption[Approximation of LeNet]{Approximation of LeNet.}
\label{fig:lenet_approximations}
\end{figure}

\begin{figure}[H]
\includegraphics[width=0.9\linewidth]{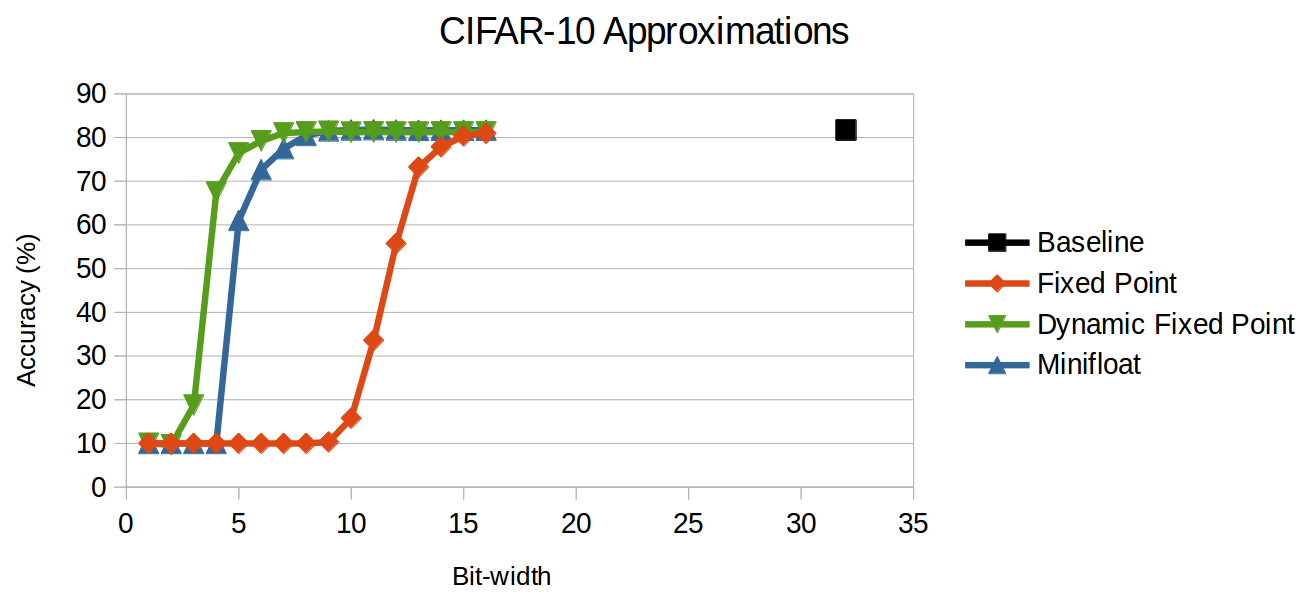}
\centering
\caption[Approximation of CIFAR-10]{Approximation of CIFAR-10.}
\label{fig:cifar10_approximations}
\end{figure}

\begin{figure}[H]
\includegraphics[width=0.9\linewidth]{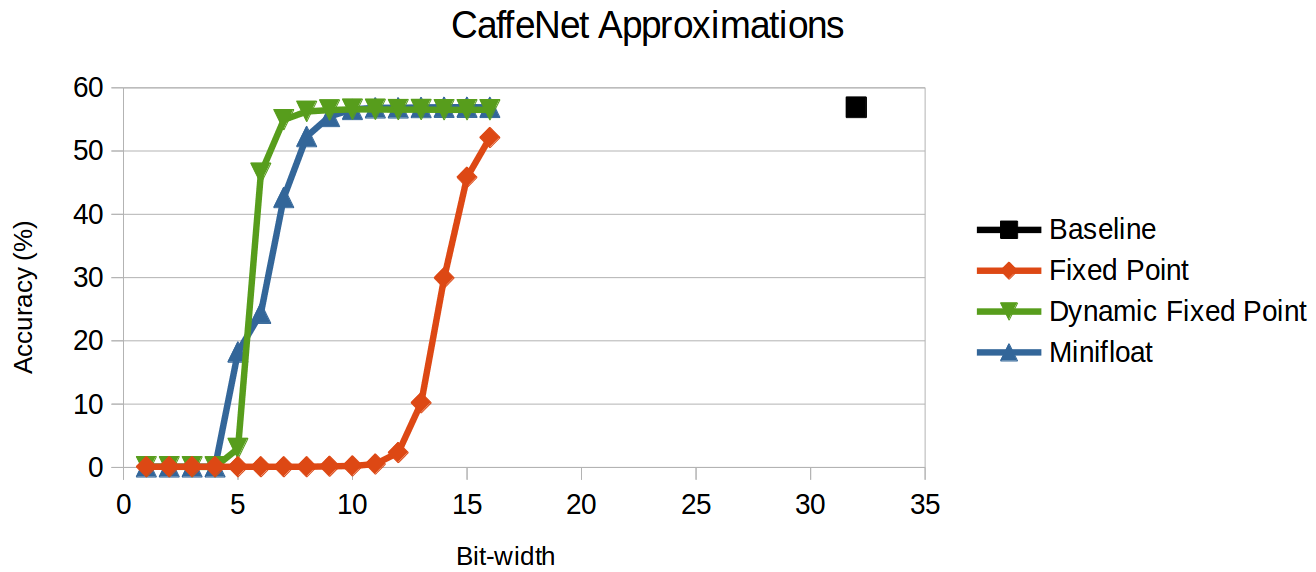}
\centering
\caption[Approximation of CaffeNet]{Approximation of CaffeNet.}
\label{fig:caffenet_approximations}
\end{figure}

In this chapter, we compare the different approximation strategies for convolutional neural
networks. For this purpose, we consider three networks: LeNet, CIFAR-10 and CaffeNet.
We analyze how well the
approximation schemes can lower the bit-width without hurting accuracy. In all
experiments, the parameters and layer outputs of convolutional and fully connected layers
are condensed to smaller bit-width. The approximation results without fine-tuning are
shown in Figures \ref{fig:lenet_approximations}, \ref{fig:cifar10_approximations},
\ref{fig:caffenet_approximations}.
For all three neural networks, dynamic fixed point has the best performance, followed by
minifloat approximation. All approximation schemes perform well at 16-bit, but as we lower
the bit-width the accuracy drops.

\section{Fixed Point Approximation}

Fixed point is the approximation scheme that requires the least energy and development
time for a hardware accelerator. However, it is also the approximation with the poorest
performance for small bit-widths. For CaffeNet for example, the dynamic range of values
is significant, as shown in subsection \ref{chap:range_caffenet}. In 15-bit convolutional
layers, 0.45\% of layer
outputs are too large and need to be represented in saturated format. When moving to 14-
bit fixed point, 2.82\% of the layer outputs are saturated. Since large layer outputs are very
important for the network's accuracy, this leads to a significant accuracy drop (see
Figure \ref{fig:lenet_approximations}).

\section{Dynamic Fixed Point Approximation}

Dynamic fixed point shows the best performance among the three approximation
schemes. Dynamic fixed point combines the advantages of both fixed point and minifloat:
On one hand this format allows to use all bits for the mantissa part which
helps for a good accuracy. On the other hand dynamic fixed point can cover a large
dynamic range, just like floating point, thanks to the exponent that is stored
implicitly. LeNet can be approximated with just 5-bit numbers, achieving the same accuracy
as the 32-bit floating point model. The same holds true for an 8-bit CIFAR-10 network.
Finally the CaffeNet architecture can be approximated with 8-bit dynamic fixed point, at an
absolute accuracy drop of 0.3\%.

\section{Minifloat Approximation}

Floating point numbers can cover a large dynamic range, thanks to their exponent. Minifloat
performs significantly better than static fixed point. However minifloat
approximation shows a sharp accuracy drop when going to very low bit-widths. This sharp drop
is at the point where there are not enough bits for the exponent. For LeNet for example, the
accuracy of 5-bit arithmetic is 97.96\%. In this setting, we use 4 bits for the exponent, no
mantissa bits and one sign bit. When we lower the bit-width further, the exponent is unable to
cover the dynamic range of values, and the accuracy drops sharply to 10.09\%.
For the other two networks, we can see a similar effect. Both CIFAR-10 and CaffeNet need 5 exponent
bits, according to Equation \ref{eq:minifloat_exp_bits}. Since we need one more bit for the sign,
those two networks need at least 6-bit minifloat numbers in order to achieve good classification
performance.

\section{Summary}

Dynamic Fixed point is very well suited for approximation of neural networks. This
approximation shows the best accuracy at low bit-widths. Although dynamic fixed point requires
some more chip area than pure fixed point arithmetic, this approximation is very well
suited for hardware acceleration of neural networks. The bit-width can be reduced to 4-bit or
8-bit for LeNet, CIFAR-10, and CaffeNet. This reduces the required
memory bandwidth and footprint significantly, which is expected to yield significant
energy savings for FPGA and ASIC designs.
\chapter{Ristretto: An Approximation Framework for Deep CNNs}

\section{From Caffe to Ristretto}

According to Wikipedia, Ristretto is `a short shot of espresso coffee made with the
normal amount of ground coffee but extracted with about half the amount of water'.
Similarly, our compressor removes the unnecessary parts of a CNN, while making sure
the essence - the ability to predict classes from images - is preserved. With its strong
community and fast training for deep CNNs, Caffe created by \cite{jia2014caffe} is an excellent
framework
to build
on. Ristretto takes a trained model as input, and automatically brews a condensed
network version. Input and output of Ristretto are a network description file (prototxt)
and the network parameters. The condensed model in Caffe-format can then be used for a
hardware accelerator.

\section{Quantization Flow}

\begin{figure}[H]
\includegraphics[width=0.99\linewidth]{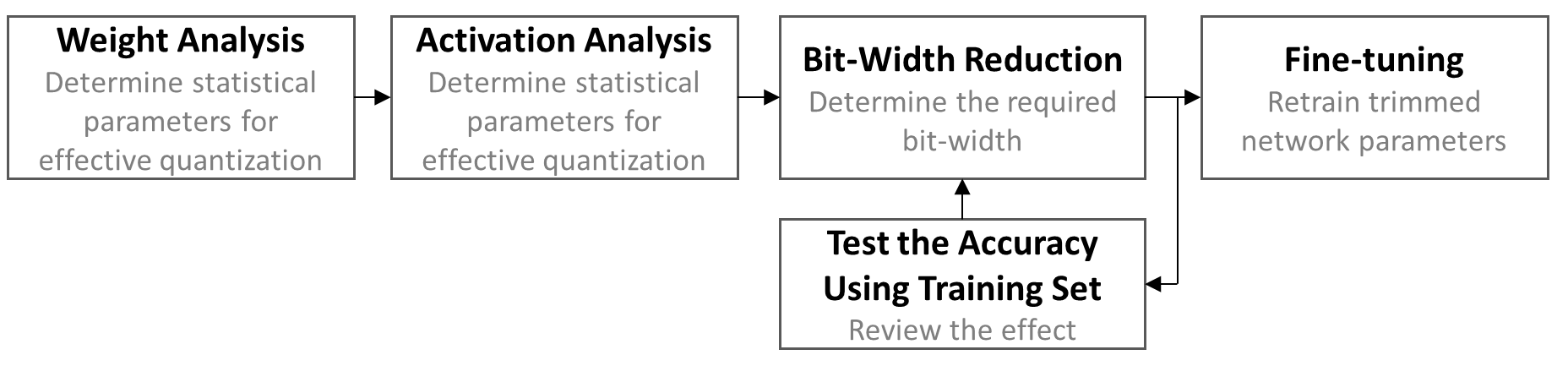}
\centering
\caption[Quantization flow]{Network approximation flow with Ristretto.}
\label{fig:quantization_flow}
\end{figure}

Ristretto can condense any 32-bit floating point network to either fixed point, minifloat
or integer power of two parameters. Ristretto's quantization flow has five stages
(Figure \ref{fig:quantization_flow}). In the first step, the dynamic range of the weights is
analyzed to find a
compressed number representation. For dynamic fixed point, Ristretto allocates enough bits to
the integer part to avoids saturation of large values.
Similarly, for minifloat approximation, the framework makes sure enough bits are allocated to
the exponent. To quantize full precision numbers into a smaller number format, Ristretto uses
round nearest even.

The second step runs several thousand images in forward path. The generated layer
activations are analyzed to generate statistical parameters. Ristretto allocates enough bits
to the new number format to avoid saturation of layer activations.

Next Ristretto performs a binary search to find the optimal number of bits for
convolutional weights, fully connected weights, and layer outputs. For dynamic fixed
point, a certain network part is quantized, while the rest remains in floating point. Since
there are three network parts that should use independent bit-widths, iteratively quantizing
one network part allows us to find the optimal bit-width for each part. Once a good
trade-off between small number representation and classification accuracy is found, the
resulting network can be fine-tuned.

\section{Fine-tuning}
\label{chap:fine-tuning}

In order to make up for the accuracy drop incurred by quantization, the quantized
network is fine-tuned in Ristretto. During this retraining procedure, the network learns
how to classify images with discrete-valued parameters $w'$. During fine-tuning, we will calculate
small weight updates $\Delta w$. Since these small weight updates may be below the quantization
step size of the discrete parameters, we also keep a set of full precision weights $w$.

\begin{figure}[H]
\includegraphics[width=0.9\linewidth]{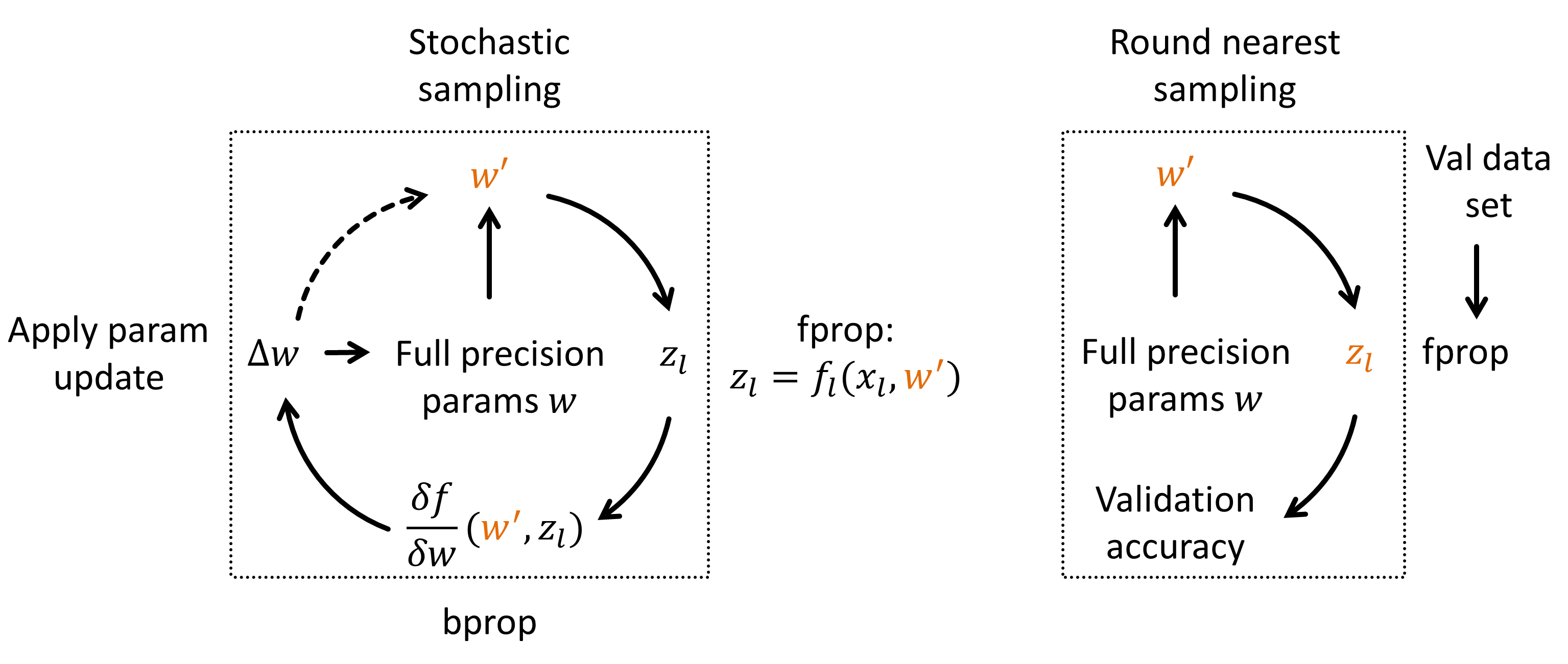}
\centering
\caption[Fine-tuning with full precision weights]{Fine-tuning with shadow weights.
The left side shows the training
process with full-precision shadow weights. On the right side the fine-tuned network is
benchmarked on the validation data set. Quantized values are represented in orange.}
\label{fig:shadow_weights}
\end{figure}

Ristretto uses the fine-tuning procedure shown in Figure \ref{fig:shadow_weights}. For each batch,
the full
precision weights are quantized to reduced-precision format. During forward
propagation, these discrete weights are used to compute the layer outputs $z_l$ . Each layer $l$
turns its input batch $x_l$ into output $z_l$, according to its function
$f_l: (x_l,w') \rightarrow z_l$. Assuming
the last layer computes the loss, we denote $f$ as the overall CNN function. The goal of
back propagation is to compute the error gradient $\frac{\delta f}{\delta w}$ with respect to
each quantized
parameter. For parameter updates we use the Adam rule by \cite{kingma2014adam}. As an important
observation, we do not quantize layer outputs during fine-tuning. We use floating point
layer outputs instead, which enables Ristretto to analytically compute the error gradient
with respect to each parameter. In contrast, scoring of the network is done with
reduced precision layer outputs.

To achieve the best fine-tuning results, we used a learning rate that is an order of
magnitude lower than the last full precision training iteration. Since the choice of hyper
parameters for retraining is crucial \citep{bergstra2012random}, Ristretto relies on minimal
human intervention
in this step.

\section{Fast Forward and Backward Propagation}

Ristretto brews a condensed network with reduced precision weights and layer
activations. For simulation of the forward propagation in hardware, Ristretto uses full
floating point for accumulation. This follows the thought of \cite{gupta2015deep} and is
conform with our
description of the forward data path in subsection \ref{chap:approximation_ristretto}.
During fine-tuning, the full precision
weights need to be quantized for each batch, but after that all computation can be done in
floating point (Figure \ref{fig:shadow_weights}). Therefore Ristretto can fully leverage
optimized matrix-
matrix multiplication routines for both forward and backward propagation. Thanks to its
fast implementation on the GPU, a fixed point CaffeNet can be tested on the ILSVRC
2014 validation dataset (50k images) in less than 2 minutes (using one Tesla K-40 GPU).

\section{Ristretto From a User Perspective}

Ristretto is based on the highly optimized Caffe-framework and follows its principles. A Caffe
user will appreciate the smooth and easy-to-understand integration of Ristretto.

\paragraph{Caffe:}

\begin{figure}[H]
\includegraphics[width=0.4\linewidth]{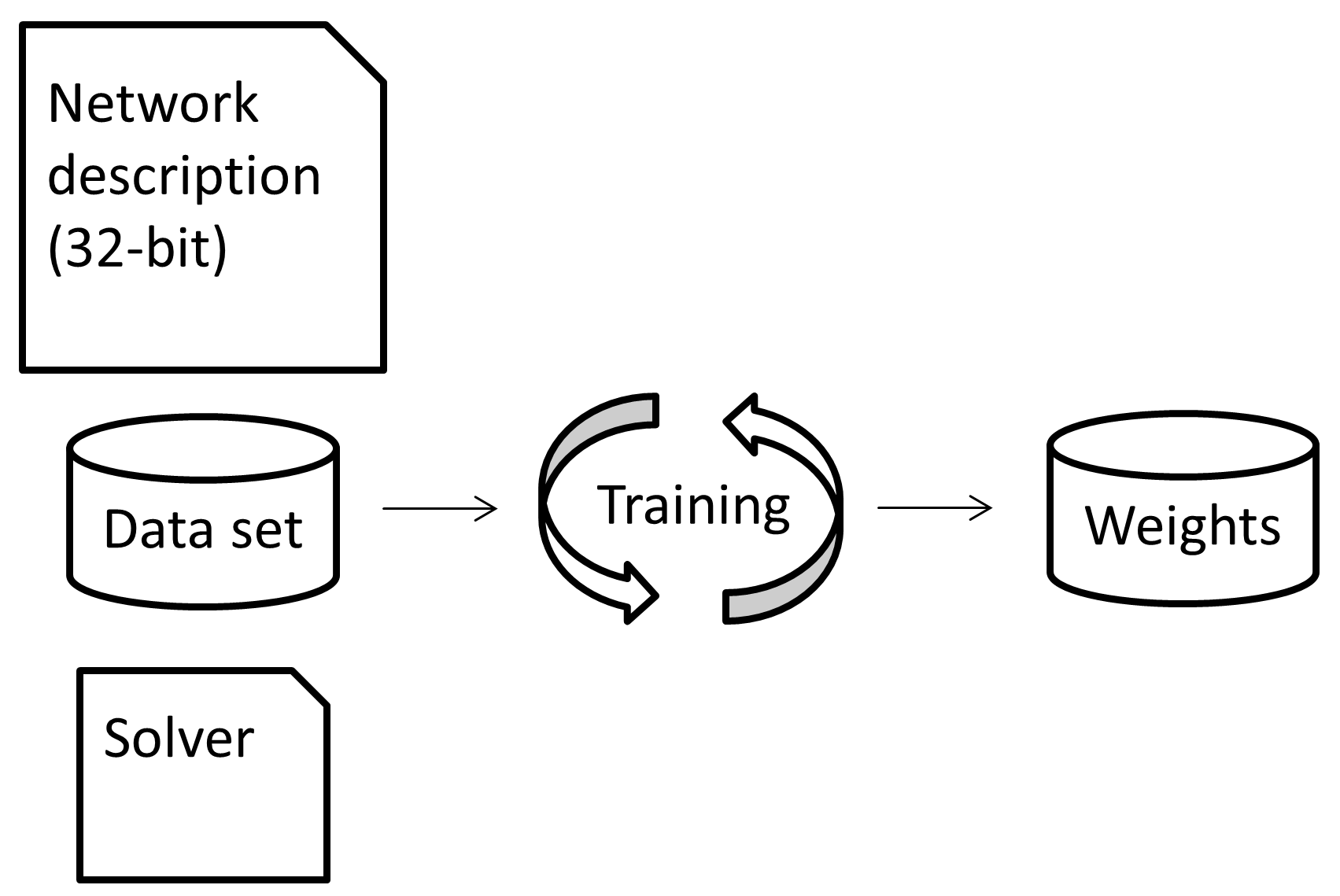}
\centering
\caption[Network brewing with Caffe]{Network training with Caffe, the base framework of
Ristretto.}
\label{fig:caffe_pipeline}
\end{figure}

Development of an image processing algorithm in Caffe starts with a network description file (see
Figure \ref{fig:caffe_pipeline}). This
file is written by the user and contains the hyper parameters of the neural network architecture.
The Caffe framework
uses the Google Protocol Buffer format to encode the network information. Networks are
represented as directed acyclic graphs. The vertices are layers which do computation based on
the input data and layer parameters. Data flows from one layer to another in so called `blobs'.

As second item, the Caffe user needs a labeled training data set.
Finally Caffe requires a 
solver file, which contains the hyper parameters for training, such as initial learning rate
and training duration.

Once the network description file, the data set and the solver are prepared, the Caffe tool
can be used for training. The result of training is a file containing the trained network parameters.
This parameter file -- together with the network description file -- can be deployed for
classification of arbitrary images.

\paragraph{Ristretto:}

\begin{figure}[H]
\includegraphics[width=0.6\linewidth]{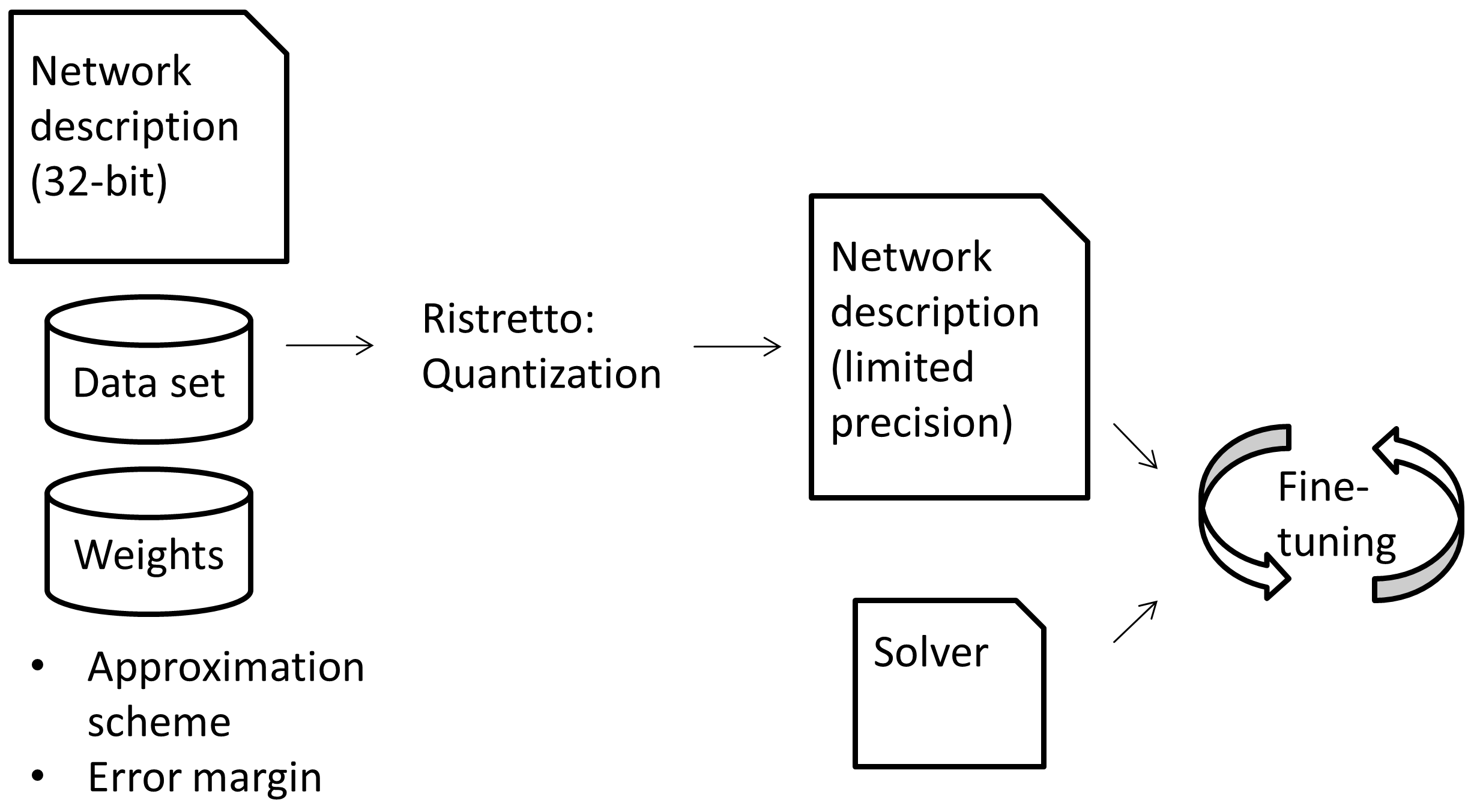}
\centering
\caption[Network quantization with Ristretto]{Network quantization and fine-tuning with Ristretto.}
\label{fig:ristretto_pipeline}
\end{figure}

Our approximation framework can be seen as a feature extension to Caffe. In fact, for the
network compression pipeline described in Figure \ref{fig:ristretto_pipeline}, some steps
require the traditional Caffe tool (which was left as-is). Ristretto starts where Caffe ends:
a trained network serves as input to the quantization pipeline. The user has several options
to choose from for quantization, for instance he can set the error margin and an approximation
strategy. The Ristretto tool quantizes the trained 32-bit floating point network to the
smallest possible bit-width representation. Ristretto produces the network description
of the condensed model, which follows the format of Caffe. The condensed model contains limited-precision layers,
which are a Ristretto-specific feature. Moreover each layer has quantization parameters, such as
the number of integer and fractional bits in the case of fixed point approximation.

At this point,
the quantized model description could be used to score the network on a data set. However, in
order to increase accuracy, the user is advised to fine-tune the new model.
The user writes a new solver file which will be used for fine-tuning. The Caffe
tool will fine-tune the condensed model to achieve the highest possible accuracy.

\section{Release of Ristretto}

Ristretto is released as open source project and has the following strengths:
\begin{itemize}
\singlespacing
  \item \textbf{Automation}: Ristretto performs automatic trimming of any given CNN.
  \item \textbf{Flexibility}: Various trimming schemes are supported.
  \item \textbf{Accuracy}: Ristretto fine-tunes trimmed networks.
  \item \textbf{Speed}: Ristretto runs on the GPU and leverages optimized CUDA routines.
\end{itemize} 
Ristretto has a homepage\footnote{\url{http://ristretto.lepsucd.com/}} and the source code is
available\footnote{\url{https://github.com/pmgysel/caffe}}.

\section{Future Work}

Ristretto follows the modular source code architecture of Caffe. New features such as
new limited precision layer types can be added to Ristretto easily. In this section
we discuss different possible future steps.

\subsection{Network Pruning}

The most energy-costly operation for CNN accelerators is off-chip memory access. Since
large networks don't fit into on-chip memory, it is imperative to compress the network.
Most network weights come from fully connected layers. It has been shown that a
significant part of the connections in fully connected layers can be removed. Previous
work \citep{han2015deep} achieves high network compression rates with no loss in classification
accuracy.

\subsection{Binary Networks}

The first published work to represent ImageNet networks with binary weights was by
\cite{rastegari2016xnor}. Their results show that very deep networks can be approximated
with binary weights, although at an accuracy drop of around 3\% for CaffeNet and 6\% for
GoogLeNet. Substituting 32-bit floating point parameters with just one bit of information
necessarily reduces the network's accuracy to extract features from images. The challenge with
binary parameters is to achieve high prediction accuracy without increasing the parameter size
or adding additional computational overhead.

\subsection{C-Code Generation}

Many academic projects use high-level synthesis tools for FPGA and ASIC based accelerators. 
The standard development tool-chain starts with a C-implementation of the algorithm, which then
undergoes many unit tests to verify correct functionality. In a next step the C-code is manually
converted to System-C which serves as input to high level synthesis (HLS). Highly optimized HLS
tools can produce very efficient Verilog code within a fraction of the time which would be
needed for manual Verilog coding. We plan to add a feature to Ristretto which allows for
automatic generation of the C-code of a condensed network. This feature will produce
the necessary code files as well as a dump of the extracted low-precision parameters.

We hope that Ristretto will enable researchers to speed up their development time
for CNN accelerators. Hopefully Ristretto will gain traction in the community. As it is an
open-source project, the community can help adding new features to the framework, which will
make the tool even more powerful.

\bibliography{mycitations}

\end{document}